\documentclass[11pt]{article}

\usepackage[final]{acl}

\usepackage{times}
\usepackage{latexsym}
\usepackage{subcaption}
\usepackage{makecell}
\usepackage[T1]{fontenc}

\usepackage[utf8]{inputenc}

\usepackage{microtype}

\usepackage{inconsolata}
\usepackage[dvipsnames]{xcolor}

\usepackage{graphicx}
\usepackage{xcolor}
\usepackage{multirow}
\usepackage{amsmath}
\usepackage{amssymb}
\usepackage{booktabs}        
\usepackage[table]{xcolor}   
\usepackage{array}  
\newcommand{\res}[2]{#1\rlap{\textsuperscript{\textbf{\color{red!50!black}\scriptsize{+#2}}}}}

\title{\texttt{LLMSurgeon}: Diagnosing Data Mixture of Large Language Models}

\author{%
  \textbf{Yaxin Luo}$^{1\ast}$,~
  \textbf{Jiacheng Cui}$^{1\ast}$,~
  \textbf{Xiaohan Zhao}$^{1}$,~
  \textbf{Xinyi Shang}$^{1,2}$,\\
  \textbf{Jiacheng Liu}$^{1}$,~
  \textbf{Xinyue Bi}$^{1}$,~
  \textbf{Zhaoyi Li}$^{1}$,~
  \textbf{Zhiqiang Shen}$^{1\dagger}$\\
  $^1$VILA Lab, MBZUAI \qquad $^2$UCL\\
  $^\ast$Equal Contribution \qquad $^\dagger$Corresponding Author\\[2mm]
  \textbf{Code \& Data:} \href{https://github.com/Yaxin9Luo/LLMSurgeon}{\textcolor{magenta}{LLMSurgeon}}
}

\begin{document}
\maketitle

\thispagestyle{plain}
\pagestyle{plain}

\begin{abstract}
The pretraining data mixture of Large Language Models (LLMs) constitutes their "digital DNA", shaping model behaviors, capabilities, and failure modes. Yet this composition is rarely disclosed, making post-hoc auditing of data combination or provenance difficult. In this work, we formalize \textbf{\texttt{Data Mixture Surgery (DMS)}}: given only generated text from a target LLM, estimate the domain-level distribution of its pretraining corpus under a predefined taxonomy. We propose \textbf{\texttt{LLMSurgeon}}, a strong framework that casts DMS as an inverse problem under the label-shift assumption\footnote{Label shift assumes that domain proportions change, while domain-specific language patterns remain unchanged.}. Rather than directly aggregating classifier outputs, \texttt{LLMSurgeon} estimates a calibrated \textit{soft} confusion matrix and solves a constrained inverse problem to correct systematic domain confusion and recover the latent mixture prior. To evaluate, we introduce \textbf{\texttt{LLMScan}}, a recipe-verifiable evaluation suite built from open-source LLMs with transparent pretraining mixtures. Across LLMScan, \texttt{LLMSurgeon} recovers domain mixtures with high fidelity under fixed protocols. Our work presents a practical, post-hoc approach for auditing the \textit{digital DNA} of foundation models without access to their training data.

\end{abstract}

\section{Introduction}

\begin{figure}[t] 
    \centering
    \includegraphics[width=\linewidth]{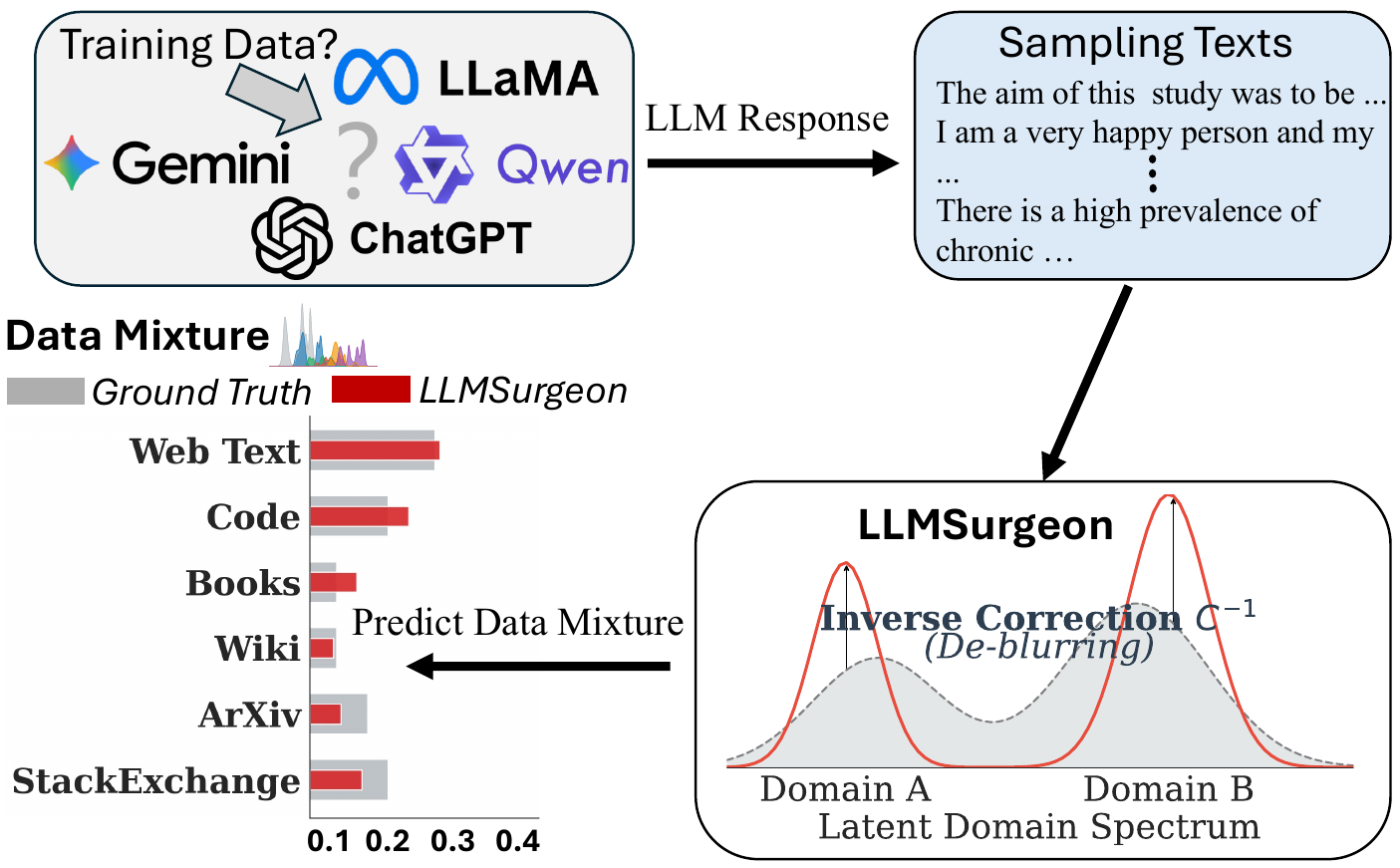}
    \caption{Overview of Data Mixture Surgery problem and the LLMSurgeon framework for solving it.}
    \label{overview}
    \vspace{-0.7em}
\end{figure}

Modern Large Language Models (LLMs)~\citep{gpt5.2,gemini3,qwen3,deepseek3.2} operate as \textit{digital alchemy}: while their capabilities in reasoning and coding are undeniable, the ingredients of their massive training corpora remain one of the most significantly guarded secrets in AI. This lack of transparency creates a critical bottleneck for safety, accountability, and governance. Without access to the \textit{digital DNA} of these models and their pretraining data composition, it is impossible to audit them for demographic biases, assess copyright infringement risks, or explain performance disparities across domains. As LLMs become central to social infrastructure, the inability to answer the simple question "\textbf{What was this LLM trained on?}" poses a fundamental challenge to LLM transparency and trustworthy AI.

To date, efforts to peer inside these black boxes have primarily relied on Membership Inference Attacks (MIA) \citep{shokri2017membership,carlini2021extracting,mink,minkpp,dcpdd}. 
While these tools are effective at the microscopic level of identifying whether a specific document was seen during training, they fail to answer macroscopic questions about data distribution.
We face a paradox: MIA tools can detect a single grain of sand (a specific sample), but lack the capacity to describe the landscape of the beach (the domain composition). Attempting to estimate global composition by aggregating millions of noisy, instance-level MIA predictions is computationally prohibitive and prone to catastrophic error accumulation.
\begin{figure*}[t] 
    \centering
    \includegraphics[width=\linewidth]{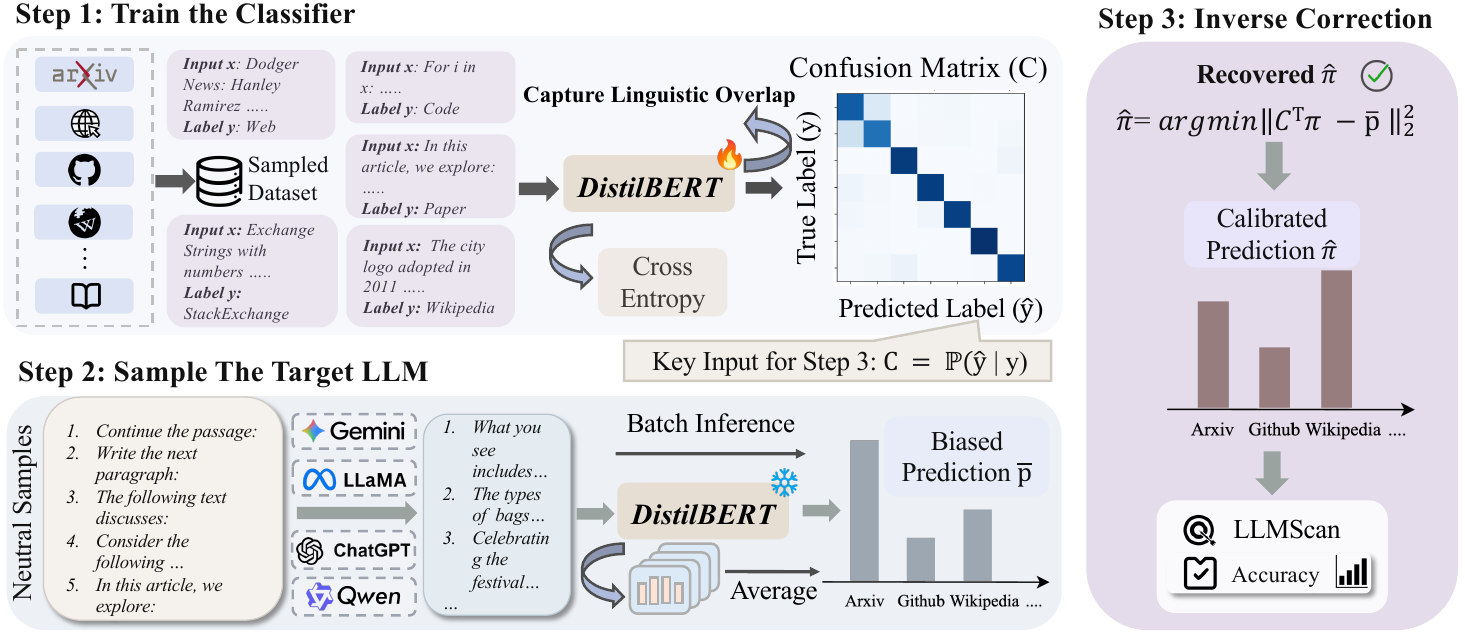}
    \caption{\textbf{Overview of our proposed LLMSurgeon framework to address the Data Mixture Surgery problem.} This figure illustrates the detailed pipeline of estimating pretraining data mixture from generated text.}
    \label{pipeline}
    \vspace{-.5em}
\end{figure*}

This gap necessitates a shift from instance-level detection to \textbf{\texttt{Data Mixture Surgery (DMS)}}: estimating the global proportions of data domains that constitute a model’s world knowledge base. Crucially, we frame DMS as a targeted auditing task rather than open-ended discovery. 
To solve this issue, we propose \textbf{\texttt{LLMSurgeon}}, a carefully designed framework that treats \texttt{DMS} as an inverse problem governed by the Label Shift hypothesis. 
In the context of unsupervised LLMs, this hypothesis posits that while the global mixture proportions of domains (the prior) shift between the original training set and the model's generated output, the linguistic characteristics defining each domain (the conditional distribution) remain statistically invariant. While prompts and alignment are known to distort generation distributions~\citep{xiao2024any,kirk2023understanding}, we employ neutral sampling to minimize such stylistic shifts, ensuring the recovered mixture faithfully reflects the pretraining prior.

Leveraging this stability, \textbf{\texttt{LLMSurgeon}} operates in a rigorous three-stage pipeline.
\textbf{First}, we pre-train an external classification model on known reference data and compute a calibrated ``soft'' confusion matrix to characterize its systematic bias between similar text domains.
\textbf{Second}, we sample the neutral output responses of target LLM and classify its generated texts using the frozen classifier, obtaining a biased observation of the latent prior.
\textbf{Finally}, unlike naive approaches that accept these biased counts directly, \texttt{LLMSurgeon} utilizes the pre-computed confusion matrix to mathematically ``de-blur'' the observations, solving the inverse problem to recover the ground-truth training proportions with high fidelity.

To rigorously evaluate \texttt{Data Mixture Surgery}, we introduce \texttt{\textbf{LLMScan}}, the first benchmark comprising open-source LLMs~\citep{olmo1,pythia,llama1,amber} with transparent, ground-truth data recipes. This prevents the common pitfall of evaluating auditing tools on synthetic data that does not reflect real-world pretraining dynamics~\citep{duan2024membership}. Experiments on \texttt{LLMScan} demonstrate that \texttt{LLMSurgeon} significantly outperforms aggregation-based baselines, offering a practical, post-hoc method for auditing foundation models without accessing their weights or training data.

Our contributions are threefold. 
(1) We formalize the \textbf{\texttt{Data Mixture Surgery (DMS)}} problem, shifting the focus from membership inference to distribution recovery.
(2) We propose \textbf{\texttt{LLMSurgeon}}, a lightweight method that accurately infers latent domain priors from generated text. 
(3) We present \textbf{\texttt{LLMScan}}, a verifiable benchmark to standardize the evaluation of \texttt{Data Mixture Surgery}.

\section{Related Work}
\noindent\textbf{Membership Inference Attack} asks whether a particular sample was included in a model’s training data~\citep{shokri2017membership}, typically by exploiting behavioral differences between training and non-training examples~\citep{shokri2017membership,carlini2021extracting}. 
Membership Inference Attack (MIA) has become standard for auditing privacy leakage and evaluating defenses such as differential privacy~\citep{abadi2016deep}, and has been extended from vision to language settings, including pretraining-corpus detection for LLMs~\citep{dcpdd,mink,minkpp,recall,Neighbor}. Despite their utility, MIA is fundamentally instance-level, it delivers a binary or probabilistic decision per example, rather than a corpus-level estimate of how training data is composed. 
MIA often relies on confidence scores, logits, or carefully matched reference datasets and can be brittle under distribution shift, calibration errors, or regularization~\citep{watson2022importance}. 
For LLMs, many techniques operate on short spans or token windows, making them expensive to scale and ill-suited for inferring high-level category proportions of large, heterogeneous corpora. 
In contrast, we pursue a \emph{corpus-level} objective: \texttt{Data Mixture Surgery}. 

\noindent\textbf{Data Mixture Optimization} focuses on selecting or reweighting pretraining data to enhance model performance, employing techniques like importance sampling and gradient-based reweighting~\citep{xie2023doremi,chen2023datajuicer,xie2023dsir,chen2024skillit}. However, these methods operate \emph{pre-hoc}, requiring full access to the training loop and raw datasets. They cannot audit fixed, closed-source models where data mixture is unknown. We tackle the problem: performing post-hoc inference of effective multi-domain mixture solely from a trained model's generated outputs.

\noindent\textbf{Dataset Usage Cardinality Inference (DUCI).}
\citet{duci} estimate the proportion of a specific, known dataset used during training by aggregating debiased membership predictions. While effective for auditing a single candidate corpus, this approach strictly requires access to the underlying data. In contrast, our work addresses \texttt{Data Mixture Surgery}, recovering the global multi-domain mixture $\pi$ solely from model generations without requiring access to the training dataset, replacing pointwise aggregation with calibrated label-shift inversion.

\section{Data Mixture Surgery } 
\subsection{Problem Formulation}
Let $\mathcal{X}$ denote space of text sequences and $\mathcal{Y} = \{1, \dots, K\}$ be a set of $K$ disjoint semantic domains (e.g., \textit{Paper}, \textit{Wikipedia}, \textit{Code}). We define the conditional distribution of text within a specific domain $i$ as $p_i(x) \triangleq p(x \mid y=i)$.

During pretraining, the LLM optimizes its parameters $\theta$ over a training corpus $\mathcal{D}_{\text{train}}$. We model this corpus as a mixture distribution defined by the ground-truth mixing vector $\boldsymbol{\alpha} \in \Delta^{K-1}$:
\begin{equation}
    p_{\boldsymbol{\alpha}}(x) = \sum_{i=1}^{K} \alpha_i p_i(x)
    \label{eq:training_mixture}
\end{equation}
where $\alpha_i$ represents the true proportion of domain $i$ in the training set (the quantity we wish to audit).

When queried with neutral prompts, the trained LLM generates samples from an induced distribution $q(x)$. Due to optimization dynamics, under-fitting, or sampling temperature, the model's internal usage of domains may diverge slightly from the exact training proportions. We model this generation distribution as:
\begin{equation}
    q_{\boldsymbol{\pi}}(x) = \sum_{i=1}^{K} \pi_i p_i(x)
    \label{eq:generation_mixture}
\end{equation}
where $\boldsymbol{\pi} \in \Delta^{K-1}$ is latent effective prior, domain mixture actually encoded by model's behavior.

\noindent\textbf{The Prediction Goal.}
Our objective is \texttt{Data Mixture Surgery.} Given a set of generated samples $X_{\text{gen}} = \{x_n\}_{n=1}^N \sim q_{\boldsymbol{\pi}}(x)$, we aim to estimate the vector $\boldsymbol{\pi}$.
We assume  generative process adheres to the \textit{Label Shift} hypothesis. Specifically, while  marginal distribution of domains shifts from $\boldsymbol{\alpha}$ (training) to $\boldsymbol{\pi}$ (generation), the conditional feature distributions remain invariant:
\begin{equation}
    q(x \mid y=i) \approx p(x \mid y=i)
    \label{eq:label_shift}
\end{equation}
This implies that when  model generates Code, it statistically resembles the Code seen during training, even if the frequency of Code generation ($\pi_{\text{code}}$) differs from its training frequency ($\alpha_{\text{code}}$). This allows us to treat the recovery of $\boldsymbol{\pi}$ as a distribution matching problem, solving for mixing coefficients that best explain  observed $X_{\text{gen}}$.

\noindent\textbf{Challenges.}
A straightforward strategy for estimating data mixture is to aggregate Membership Inference Attacks (MIA) signals across sampled corpora. However, this naive approach faces three critical bottlenecks.
\textbf{Token Limitations:} MIAs are typically designed for short sequences, making them computationally prohibitive to scale across large, document-level corpora.
\textbf{Error Accumulation:} Pointwise prediction errors compound when aggregated over millions of samples, leading to high-variance distribution estimates.
\textbf{Algorithmic Bias:} MIAs often exhibit domain-dependent performance (e.g., higher accuracy on memorized code than generic text), introducing systematic skew that distorts the recovered mixture proportions.

\begin{table*}[t]
\centering
\small
\resizebox{\linewidth}{!}{
\begin{tabular}{lcccc}
\toprule
\textbf{Target Model} & \textbf{Parameters} & \textbf{Pretraining Corpus} & \textbf{Granularity Level} & \textbf{Pre-Defined Domains ($K$)} \\
\midrule
\multicolumn{5}{l}{\textit{\textbf{General Purpose Models}}} \\
LLaMA-1~\citep{llama1} & 7B, 65B & Public Mix (CC, Wiki, etc.) & Coarse-Grained & 6 \\
OLMo~\citep{olmo1} & 1B & Dolma~\citep{soldaini2024dolma} & Coarse-Grained & 6 \\
Amber~\citep{amber} & 13B & LLM360 Mix~\citep{amber} & Coarse-Grained & 6 \\
\midrule
\multicolumn{5}{l}{\textit{\textbf{Pile-Based Models}}} \\
Pythia~\citep{pythia} & 2.8B, 12B & The Pile~\citep{gao2020pile} & Mid-Grained & 17 \\
GPT-Neo~\citep{gptneo} & 2.7B & The Pile & Mid-Grained & 17 \\
\midrule
\multicolumn{5}{l}{\textit{\textbf{Domain Specialized Models}}} \\
StarCoder~\citep{starcoder} & 15.5B & The Stack~\citep{kocetkov2022stack} & Fine-Grained & 87 \\
\bottomrule
\end{tabular}
}
\caption{\textbf{Overview of the LLMScan Benchmark Suite.} The benchmark covers three levels of auditing granularity (Coarse, Mid, Fine) across varying model scales (1B to 65B).}
\vspace{-.5em}

\label{tab:becnhmark_info}
\end{table*}

\subsection{LLMScan: First Benchmark for Data Mixture Surgery}
\label{sec:llmscan}
Evaluating DMS requires ground truth that is largely absent in closed-source AI. To bridge this gap, we introduce \texttt{\textbf{LLMScan}}, a benchmark comprising 8 open-source foundation models (1B--65B parameters) with verifiable data genealogies~\citep{olmo1,llama1,amber,pythia,starcoder}. As summarized in Table~\ref{tab:becnhmark_info}, we evaluate auditing performance across three resolution levels to assess robustness against semantic overlap. The \textbf{Coarse-Grained} setting ($K=6$) utilizes SlimPajama-DC~\citep{shen2023slimpajama} definitions to audit general-purpose models (LLaMA-1, OLMo, Amber), merging overlapping web sources (e.g., C4 and CommonCrawl) for stability. The \textbf{Mid-Grained} setting ($K=17$) employs The Pile~\citep{gao2020pile} taxonomy to analyze the Pythia family and GPT-Neo, challenging the auditor to distinguish more sub-domains. Finally, the \textbf{Fine-Grained} setting ($K=87$) uses The Stack~\citep{kocetkov2022stack} to distinguish specific programming languages in StarCoder. To establish the ground truth, we define the candidate domains $K$ according to each model's official pretraining documentation, extracting the exact composition vector directly from their technical reports. By strictly adhering to documented recipes rather than synthetic mixtures, \texttt{LLMScan} ensures that auditing performance is measured against real-world, large-scale training dynamics.

\section{LLMSurgeon: A Simple Data Mixture Surgery Method}
\label{llmsurgeon}
\vspace{-0.15em}

Our framework recovers the latent effective prior $\boldsymbol{\pi}$ by treating the problem as a \textit{label-shift inversion} task. We decompose the process into three stages: (1) Characterizing the systematic bias of a proxy classifier, (2) Sampling the target LLM's outputs, and (3) Solving the constrained inverse problem.

\subsection{Characterizing Systematic Bias}
\label{sec:confusion_matrix}
Since we cannot access the target LLM's internal states, we employ an external proxy domain classifier $f_{\phi}: \mathcal{X} \to \Delta^{K-1}$. However, no classifier is perfect; applying $f_{\phi}$ directly to generated text yields a biased estimate due to domain confusion (e.g., confusing \textit{C++} with \textit{C}).

We explicitly model this error profile as a linear operator. Using a held-out reference dataset $\mathcal{D}_{\text{ref}}$ where the ground-truth domain labels are known, we compute the \textbf{soft confusion matrix} $C \in \mathbb{R}^{K \times K}$. Each entry $C_{ij}$ represents the expected probability that the classifier predicts domain $j$ given a sample truly from domain $i$:
\begin{equation}
    C_{ij} = \mathbb{E}_{x \sim p_i} \left[ f_{\phi}(x)_j \right]
    \label{eq:confusion_matrix}
\end{equation}
where $f_{\phi}(x)_j$ denotes the predicted probability for class $j$. Here, $C$ serves as a \textit{calibration operator} that maps the true domain distribution to the classifier's biased observation space. If the classifier were perfect, $C$ would be the identity matrix $I$. In reality, off-diagonal elements capture the semantic overlap between domains.

\subsection{Observing the Target Distribution}
\label{sec:sampling}
To probe the target LLM, we generate a corpus of synthetic text $X_{\text{gen}} = \{x_n\}_{n=1}^N$ using neutral prompts designed to trigger the model's natural domain prior $q_{\boldsymbol{\pi}}$ (as defined in Eq. \ref{eq:generation_mixture}).

We pass these generations through our proxy classifier to obtain the \textbf{empirical mean prediction vector} $\bar{\mathbf{p}} \in \mathbb{R}^K$:
\begin{equation}
    \bar{\mathbf{p}} = \frac{1}{N} \sum_{n=1}^{N} f_{\phi}(x_n)
    \label{eq:empirical_mean}
\end{equation}
Crucially, notice that $\bar{\mathbf{p}}$ is not the true distribution $\boldsymbol{\pi}$, but rather the convolved observation corrupted by the proxy classifier's bias.

\subsection{The Inverse Surgery: Recovering \texorpdfstring{$\boldsymbol{\pi}$}{pi}}
\label{sec:inverse_solution}
We now link observed signal $\bar{\mathbf{p}}$ to latent target $\boldsymbol{\pi}$. By linearity of expectation and the definition of generation mixture $q_{\boldsymbol{\pi}}(x) = \sum_k \pi_k p_k(x)$, expected output of classifier is:
\begin{align}
    \mathbb{E}_{x \sim q_{\boldsymbol{\pi}}} [f_{\phi}(x)] &= \sum_{k=1}^K \pi_k \underbrace{\mathbb{E}_{x \sim p_k} [f_{\phi}(x)]}_{\text{Row } k \text{ of } C} \nonumber \\
    &= C^\top \boldsymbol{\pi}
\end{align}
This derivation reveals that observed distribution $\bar{\mathbf{p}}$ approximates $C^\top \boldsymbol{\pi}$. Consequently, recovering true prior $\boldsymbol{\pi}$ becomes a constrained linear inverse problem. We solve for optimal $\hat{\boldsymbol{\pi}}$ that minimizes reconstruction error:
\begin{equation}
    \hat{\boldsymbol{\pi}} = \underset{\boldsymbol{\pi} \in \Delta^{K-1}}{\arg\min} \; \left\| C^\top \boldsymbol{\pi} - \bar{\mathbf{p}} \right\|_2^2
    \label{eq:optimization}
\end{equation}
subject to constraints $\sum \pi_k = 1$ and $\pi_k \ge 0$.

\paragraph{Why Direct Audit-by-Aggregation is Biased for DMS.}
Prior work shows that membership inference reliability depends strongly on access assumptions, calibration, input scale, and distributional conditions~\citep{maini2021dataset,carlini2021extracting,zhang2024membership,meeus2025sok,duan2024membership,chen2025statistical}. Our claim is that MIA is not weak for its native objective, directly aggregating membership-like scores does not yield a calibrated estimator of domain mixture. DMS requires recovering a simplex-valued distribution under domain-dependent confusion, so naive aggregation inherits domain bias and accumulation error. This motivates our calibrated inverse formulation.

\section{Experiments}

\subsection{Experimental Settings and Metrics}
\label{sec:exp_settings}
We evaluate \texttt{LLMSurgeon} on \texttt{LLMScan}, our proposed benchmark with publicly documented pretraining data mixtures. To assess the robustness of \texttt{LLMSurgeon} across different resolutions, we conduct experiments at three distinct granularity levels. For each level, we select a representative open-source dataset as reference corpus to train domain classifier $f_{\phi}$ and compute confusion matrix $C$. 
For the \textbf{Coarse-Grained} setting, we utilize SlimPajama-627B-DC as the reference pool, sampling 5,000 documents from each of the 6 broad data domains for classifier training. 
In the \textbf{Mid-Grained} setting, we increase resolution to 17 diverse domains\footnote{We remove 5 copyright infringement domains.} and use The Pile to capture finer distributional shifts. 
Finally, for the \textbf{Fine-Grained} setting, we focus specifically on programming domain using The Stack, distinguishing between 87 different programming languages.
Considering the correctness of the ground truth, we select a suite of fully open-source LLMs where the pretraining data mixtures are publicly available, serving as the ground truth for quantitative evaluation. Our target models include LLaMA-1~\citep{llama1}\footnote{We select LLMs where the pretraining data mixtures are publicly available to ensure reliable of results.}, OLMo~\citep{olmo1}, Amber~\citep{amber}, Pythia~\citep{pythia},  StarCoder~\citep{starcoder} and GPT-Neo~\citep{gptneo}. 
The set of candidate domains $K$ is defined according to  official pretraining documentation under a closed-world assumption. For instance, when analyzing StarCoder, we restrict the taxonomy to the 87 coding languages present in its training data. To infer the mixture, we generate outputs using neutral prompts  and apply the \texttt{LLMSurgeon} pipeline described in Section~\ref{llmsurgeon}. To rigorously quantify reconstruction fidelity, we report Overlap Accuracy ($1 - \frac{1}{2} \sum_{k=1}^{K} | \alpha_k - \hat{\pi}_k |$) as our primary metric, alongside Mean Absolute Error (MAE) for average deviation and Coefficient of Determination ($R^2$) to evaluate structural correlation. All inference experiments for LLaMA1-65B are conducted on 4 NVIDIA A100, while all other models are evaluated on NVIDIA RTX 4090.

\begin{table*}[t]
\centering
\setlength{\tabcolsep}{3pt}
\renewcommand{\arraystretch}{1.15}
\resizebox{\linewidth}{!}{
\begin{tabular}{l !{\vrule width 1.1pt} c|c|c|c|c|c|c|c}
\specialrule{1.2pt}{0pt}{0pt}
\multirow{2.5}{*}{\textbf{Methods}} & 
\multicolumn{4}{c|}{\textbf{Coarse-Grained (Easy)}} & 
\multicolumn{3}{c|}{\textbf{Mid-Grained (Middle)}} & 
\multicolumn{1}{c}{\textbf{Fine-Grained (Hard)}} \\
& \textbf{OLMo-1B \hspace{0.2em}} & \textbf{LLaMA1-7B} & \textbf{Amber-13B} & \textbf{LLaMA1-65B} & \textbf{GPT-Neo-2.7B} & \textbf{Pythia-2.8B} & \textbf{Pythia-12B} & \textbf{StarCoder-15.5B}\\
\specialrule{0.9pt}{0pt}{0pt}
\multicolumn{9}{l}{\textbf{MIA Methods:}}\\
Joint-Logit~\citep{carlini2022membership} & 35.20 & 35.03 & 41.50 & 35.05 & 52.19 & 52.23 & 52.26 & 25.60 \\
 Loss~\citep{yeom2018privacy}               & 29.74 & 33.86 & 40.14 & 34.45 & 53.54 & 53.27 & 52.36 & 25.76\\
Ref~\citep{Ref}                             & 38.20 & 35.15 & 41.52 & 35.09 & 52.33 & 52.33 & 52.33 & 25.46\\
GradNorm~\citep{wang2024pandora}            & 34.72 & 36.03 & 39.41 & 46.52 & 58.78 & 55.66 & 35.98 & 27.54\\
 Zlib~\citep{Zlib}                          & 29.30 & 28.11 & 36.73 & 38.08 & 55.16 & 53.86 & 47.80 & 21.06\\
Neighbor~\citep{Neighbor}                   & 41.74 & 40.13 & 40.31 & 35.74 & 51.85 & 52.21 & 52.90 & 25.47\\
 Min-K\%~\citep{mink}                       & 30.84 & 28.12 & 36.81 & 23.41 & 53.47 & 53.24 & 50.23 & 23.41 \\
 Min-K\%\texttt{++}~\citep{minkpp}          & 30.99 & 32.65 & 40.56 & 32.24 & 53.47 & 53.97 & 33.09 & 23.57\\
 DC-PDD~\citep{dcpdd}                       & 35.93 & 34.82 & 38.58 & 35.03 & 56.07 & 55.43 & 52.53 & 25.98 \\
 Recall~\citep{recall}                      & 48.05 & 35.08 & 41.55 & 35.08 & 49.04 & 55.23 & 52.63 & 25.91 \\
 DUCI~\citep{duci}                          & 35.16 & 35.22 & 41.30 & 35.27 & 52.86 & 52.62 & 52.62 & 25.44\\
 
\specialrule{0.7pt}{0pt}{0pt}
\multicolumn{9}{l}{\textbf{CCI Methods:}}\\
\rowcolor{gray!10} \texttt{LLMSurgeon} (Ours) & 
\res{94.46}{46.4} & 
\res{95.14}{55.0} & 
\res{78.87}{37.3} & 
\res{94.26}{47.7} & 
\res{61.86}{3.1} &
\res{63.20}{7.5} & 
\res{65.98}{13.1} & 
\res{30.37}{2.8} 
\\
\specialrule{1.2pt}{0pt}{0pt}
\end{tabular}
}
\vspace{-0.35em}
\caption{LLMScan Benchmark. The performances are reported in overlap accuracy \text{\%}.}
\label{main_results}
\vspace{-0.85em}
\end{table*}
\vspace{-0.35em}

\subsection{Baselines}

\noindent\textbf{Adapted audit-by-aggregation baselines.} 
There is currently no established baseline tailored to black-box DMS under a predefined taxonomy. We therefore compare LLMSurgeon against pragmatic audit-by-aggregation references, including adapted MIA-style scores and DUCI-style dataset-level estimators~\citep{carlini2021extracting,yeom2018privacy,wang2024pandora,recall,Neighbor,minkpp}. These baselines are not like-for-like competitors under identical assumptions; rather, they represent natural auditing heuristics a practitioner may try in the absence of a dedicated DMS method. 
MIA methods typically output a binary prediction $\hat{y}_{i}^{(c)} \in \{0,1\}$ for a single sample $i$ from domain $c$, where $1$ indicates the sample was likely seen during training. 
To convert these instance-level signals into a domain proportion estimate, we aggregate predictions over a sampled validation set ($N=5000$ per domain). We define the MIA-inferred proportion $r_c$ for domain $c$ as the normalized count of positive predictions:
\begin{equation}
    r_{c} = \frac{\sum_{i=1}^{N}\hat{y}_{i}^{(c)}}{\sum_{j=1}^{K}\sum_{i=1}^{N}\hat{y}_{i}^{(j)}}.
\end{equation}
This baseline tests whether simply counting "detected" samples from each domain can serve as a proxy for the data mixtures.

\noindent\textbf{w/o Inverse Correction baseline.}
\label{no_inverse_baseline}
To isolate the gain from our label-shift inversion, we evaluate a \textit{Direct Estimation} baseline ($\hat{\pi}_{\text{direct}} = \bar{p}$), which uses the raw aggregated classifier outputs defined in Eq.~\ref{eq:empirical_mean} without correction.
Given the high accuracy of the proxy classifier, this baseline establishes a strong performance lower bound.
This comparison serves to verify whether our mathematical calibration provides the necessary refinement to rectify systematic biases beyond naive classification.

\subsection{Results of LLMScan Benchmark} 
Table~\ref{main_results} shows that \texttt{LLMSurgeon} consistently improves over the resource-matched direct estimator and remains more stable than adapted audit-by-aggregation references, indicating that the key gain comes from calibrated inverse correction rather than additional access. On general-purpose models like LLaMA-1-7B and OLMo-1B, we achieve overlap accuracies of \textbf{95.14\%} and \textbf{94.46\%} respectively, whereas the strongest baselines (e.g., \textit{Neighbor}, \textit{Recall}) struggle to exceed 50\%. This performance remains robust across model scales (e.g., LLaMA-7B to 65B), suggesting our method effectively captures the fundamental generation probability $q(x)$. In the challenging fine-grained setting (StarCoder), semantic blurring between similar languages lowers absolute accuracy to 30.37\%, yet \texttt{LLMSurgeon} still surpasses the best baseline (\textit{GradNorm}: 27.54\%), proving that our inverse bias correction provides the most reliable estimate even when domain boundaries are indistinct.

\subsection{Ablation Studies}

\noindent\textbf{Effect of Classifier Backbone.} To assess the effect of different classifier backbones on the overall performance, we conduct an ablation study. 
Specifically, we evaluate four settings: fine-tuned DistilBERT, DistilBERT architecture transformer classifier trained from scratch, TF-IDF, and a simple MLP classifier, as shown in Table~\ref{Classifier effect}.
We observe that fine-tuned DistilBERT consistently outperforms other backbones across most settings, achieving an absolute improvement of 4.92\% over the second-best classifier on LLaMA1-7B, and a gain of 1.81\% under the more fine-grained detection setting with StarCoder. Based on these results, we adopt fine-tuned DistilBERT as the default classifier backbone in all subsequent experiments.

\begin{table}[t]
\centering
\setlength{\tabcolsep}{2.5pt}
\renewcommand{\arraystretch}{1.15}
\resizebox{\linewidth}{!}{
\begin{tabular}{l !{\vrule width 1.1pt} cccccc}
\specialrule{1.2pt}{0pt}{0pt}
\textbf{Classifier} & \textbf{OLMo-1B}  & \textbf{LLaMA1-7B} & \textbf{Amber-13B} & \textbf{StarCoder-15.5B} & \textbf{LLaMA1-65B}\\
\specialrule{0.9pt}{0pt}{0pt}
TF-IDF  & 85.07  & 86.83 & 59.61 & 28.56 & 92.64 \\
 MLP    & 74.52 & 82.97 & 64.77 & 21.57 & 85.36 \\
 Transformer     & 89.18  & 90.22 & 75.88 & 23.11 & 94.25 \\
\rowcolor{gray!10} DistilBERT    & 94.46  & 95.14 & 78.87 & 30.37 & 94.26  \\
\specialrule{1.2pt}{0pt}{0pt}
\end{tabular}
}
\vspace{-0.35em}
\caption{Classifier performance effect ablation study.}
\vspace{-0.95em}
\label{Classifier effect}
\vspace{-0.25em}
\end{table}

\noindent\textbf{Effect of Domain Granularity.}
\begin{figure*}[t] 
    \centering
    \includegraphics[width=\linewidth]{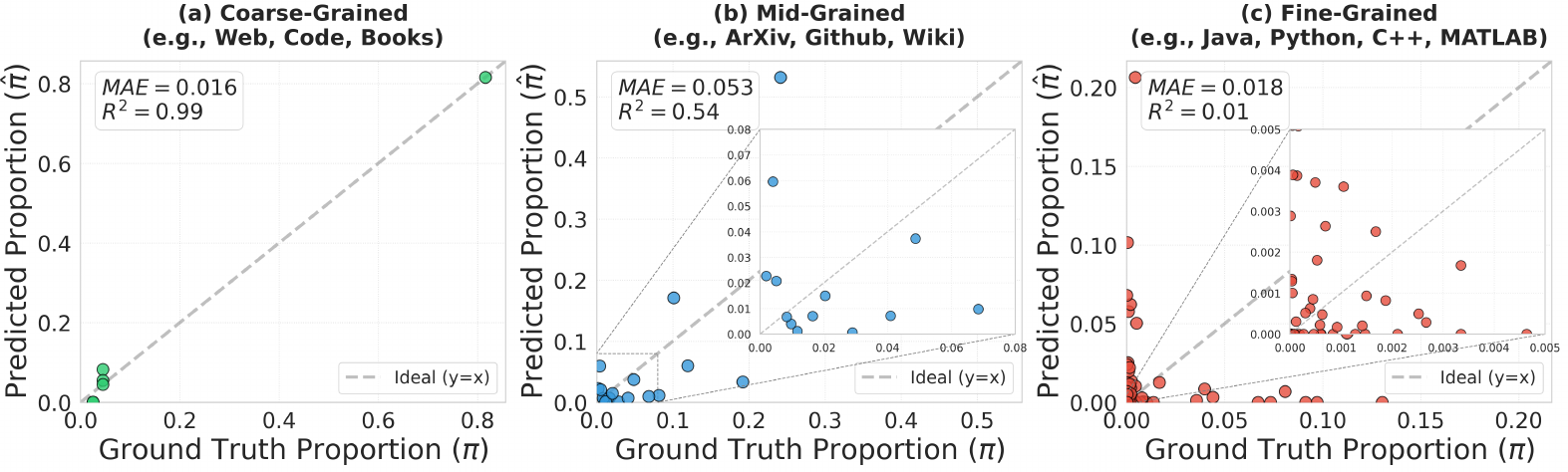}
    \caption{\textbf{Impact of Domain Granularity on Mixture Recovery.} We observe a performance hierarchy where coarse-grained recovery is near-perfect ($R^2=0.99$), whereas fine-grained estimation suffers ($R^2=0.01$) due to the high semantic confusion between similar categories (e.g., C vs. C++).}
    \label{grain}
    \vspace{-.5em}
\end{figure*}
To determine resolution limits, we evaluate \texttt{LLMSurgeon} across Coarse, Mid, and Fine granularities in Figure~\ref{grain}. Results reveal a hierarchy driven by semantic separability. Coarse-grained is near-perfect ($R^2=0.99$) due to distinct linguistic boundaries, while mid-grained estimation remains robust ($R^2=0.54$) despite increased topical overlap. In fine-grained setting, high semantic confusion (e.g., between C and C++) degrades correlation ($R^2=0.01$) by creating an ill-conditioned inverse problem. However, consistently low MAE ($0.018$) confirms that \texttt{LLMSurgeon} still successfully filters irrelevant domains, validating its utility for macroscopic auditing over microscopic dialect identification.

\begin{figure}[t] 
    \centering
    \includegraphics[width=\linewidth]{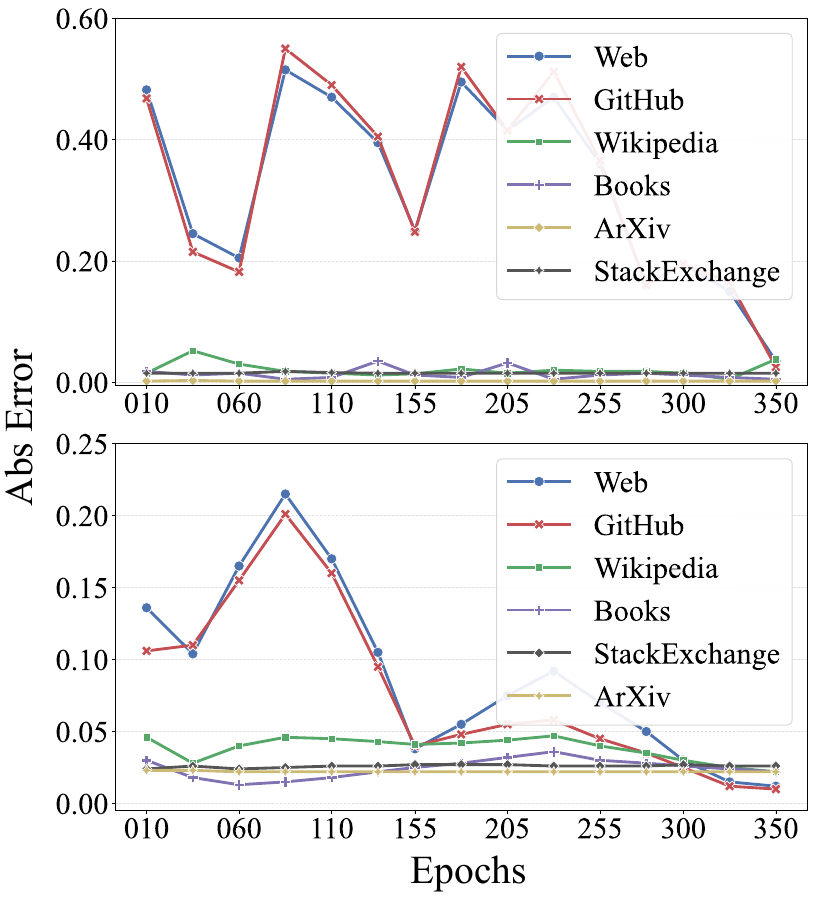}
    \caption{\textbf{Trajectories of domain estimation error throughout training.} \textbf{Top:} Amber-13B; \textbf{Bottom:} OLMo-1B. Despite different intermediate training dynamics, \texttt{LLMSurgeon} consistently converges to the ground-truth composition in the final checkpoints.}
    \label{ablation_steps}
    \vspace{-1.5em}
\end{figure}

\noindent\textbf{Effect of LLM's Pretraining Steps.}
To investigate the evolution of internal domain priors, we apply \texttt{LLMSurgeon} to intermediate checkpoints of Amber-13B and OLMo-1B. 
Figure~\ref{ablation_steps} reveals distinct training dynamics.
Amber (top) exhibits a "fluctuation-then-convergence" pattern, where dominant domains (\textit{Web, GitHub}) show high volatility during intermediate stages, likely reflecting curriculum learning or staged data injection strategies.
In contrast, OLMo (bottom) displays a significantly more stable trajectory with lower error variance, suggesting a consistent data mixing strategy throughout training.
Despite these dynamic differences, both models achieve sharp error reduction in the final phase. This confirms that \texttt{LLMSurgeon}  recovers the composition of converged models and serves as a transparent tool for monitoring training stability and data scheduling effects.

\noindent\textbf{Effect of Data Sampling Population.} We evaluate how the number of reference documents ($N$) used to train the proxy classifier $f_{\phi}$ impacts estimation fidelity as shown in Table~\ref{Training Steps}. We observe that small sample sizes ($N=100$) result in poor generalization (e.g., 20.15\% on StarCoder). Increasing $N$ to 1,000 yields substantial gains ($>10\%$ on average), but performance saturates at $N=5,000$. Interestingly, further increasing $N$ to 10,000 offers negligible improvement or even slight regression (e.g., on LLaMA-7B), likely due to redundant or noisy samples interfering with the classifier. This suggests that $N=5,000$ sufficiently captures domain features without introducing excessive noise or computational overhead. Consequently, we adopt $N=5,000$ as the optimal trade-off between accuracy and computational efficiency.

\begin{figure}[t] 
    \centering
    \includegraphics[width=\linewidth]{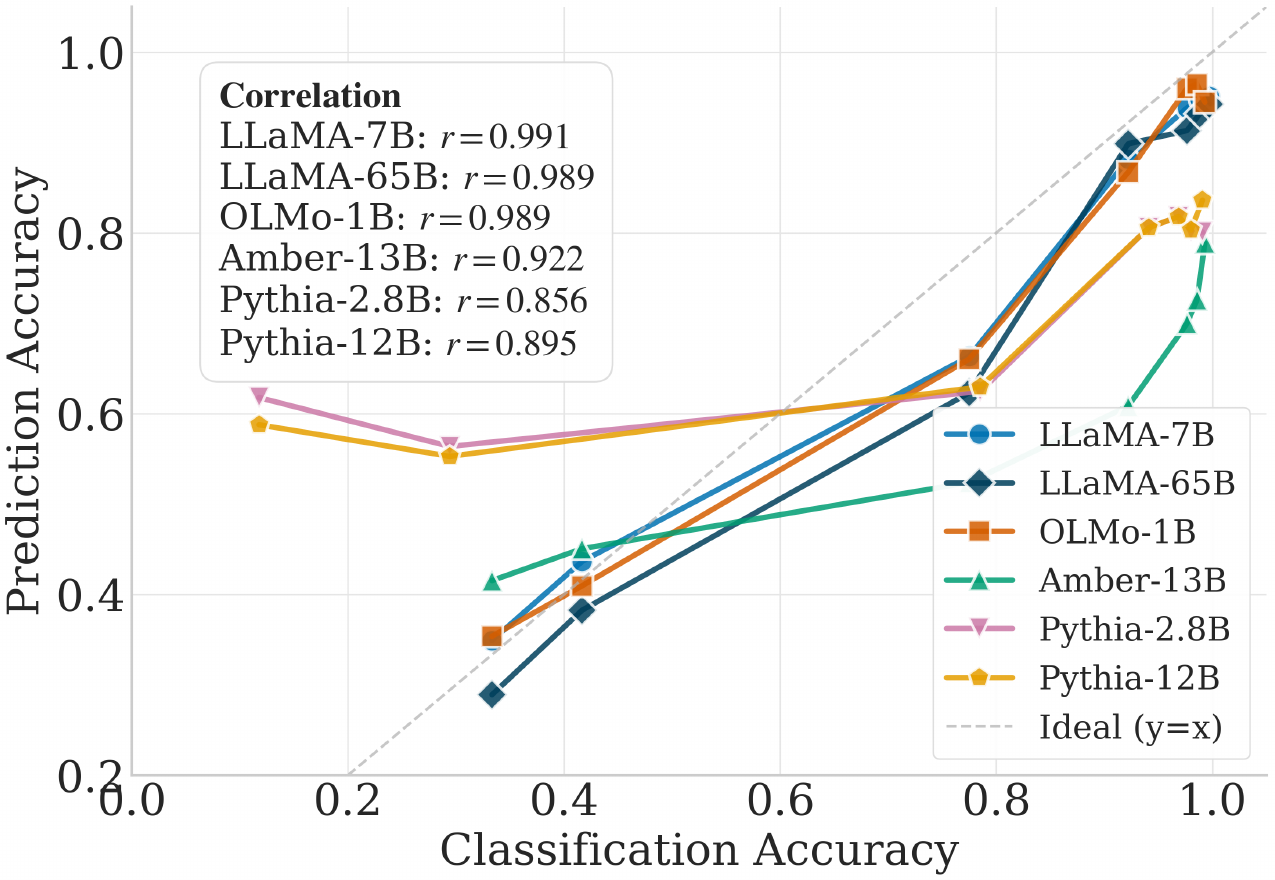}
    \vspace{-1.7em}
    \caption{\textbf{Classification Acc vs Estimation Acc.} We observe a strong positive correlation (avg $r > 0.9$) between the proxy classifier's performance and the final mixture recovery accuracy.}
    \vspace{-.5em}
    \label{class_vs_est}
\end{figure}

\begin{table}[t]
\centering
\setlength{\tabcolsep}{2.5pt}
\renewcommand{\arraystretch}{1.15}
\resizebox{\linewidth}{!}{
\begin{tabular}{c !{\vrule width 1.1pt} cccccc}
\specialrule{1.2pt}{0pt}{0pt}
\textbf{Samples/Domain} & \textbf{OLMo-1B}  & \textbf{LLaMA1-7B} & \textbf{Amber-13B} & \textbf{StarCoder-15.5B} & \textbf{LLaMA1-65B}\\
\specialrule{0.9pt}{0pt}{0pt}
100 & 73.28  & 85.78 & 71.77 & 20.15 & 83.77 \\
1000 & 95.91  & 93.68 & 74.72 & 25.62 & 93.88 \\
\rowcolor{gray!10} 5000 & 94.46  & 95.14 & 78.87 & 30.37 & 94.26 \\
10000 & 93.98 & 92.44 & 82.83 & 29.51 & 93.78 \\

\specialrule{1.2pt}{0pt}{0pt}
\end{tabular}
}
\vspace{-0.1in}
\caption{Effect of Data Sampling Population.}
\label{Training Steps}
\vspace{-.8em}
\end{table}

\begin{table}[t]
\centering
\setlength{\tabcolsep}{2.5pt}
\renewcommand{\arraystretch}{1.15}
\resizebox{\linewidth}{!}{
\begin{tabular}{l !{\vrule width 1.1pt} cccccc}
\specialrule{1.2pt}{0pt}{0pt}
\textbf{Style} & \textbf{OLMo-1B} & \textbf{LLaMA1-7B} & \textbf{Amber-13B} & \textbf{StarCoder-15.5B} & \textbf{LLaMA1-65B}\\
\specialrule{0.9pt}{0pt}{0pt}
Instructional    & 80.50 & 89.38 & 70.34 & 24.83 & 90.69 \\
Expository       & 22.71  & 90.60 & 91.53 & 25.62 & 90.93 \\
Conversational   & 85.18  & 85.70 & 92.35 & 28.00 & 86.99 \\
Coding           & 22.72  & 66.65 & 36.78 & 27.83 & 55.13 \\
Math             & 79.48  & 79.56 & 67.15 & 24.16 & 78.97 \\
\rowcolor{gray!10} Neutral          & 94.46  & 95.14 & 78.87 & 30.37 & 94.26 \\
\specialrule{1.2pt}{0pt}{0pt}
\end{tabular}
}
\vspace{-0.35em}
\caption{Robustness of Sampling Text.}
\label{Prompts Style}
\vspace{-0.2em}
\end{table}

\begin{table}[t]
\centering
\setlength{\tabcolsep}{2.5pt}
\renewcommand{\arraystretch}{1.15}
\begin{subtable}{\linewidth}
    \centering
    \resizebox{\linewidth}{!}{
    \begin{tabular}{l !{\vrule width 1.1pt} ccccc}
    \specialrule{1.2pt}{0pt}{0pt}
    \textbf{Method} & \textbf{OLMo-1B} & \textbf{LLaMA1-7B} & \textbf{Amber-13B} & \textbf{StarCoder-15.5B} & \textbf{LLaMA1-65B} \\
    \specialrule{0.9pt}{0pt}{0pt}
    w/o Inverse Correction & 92.77 & 93.42 & 77.38 & 26.47 & 93.38 \\
    \rowcolor{gray!10} \texttt{LLMSurgeon}  & 94.46 & 95.14 & 78.87 & 30.37 & 94.26 \\
    \specialrule{1.2pt}{0pt}{0pt}
    \end{tabular}
    }
    \vspace{-0.25em}
    \caption{ Importance of Inverse Correction}
    \label{tab:classifier_comp}
\end{subtable}

\vspace{1em} 

\begin{subtable}{\linewidth}
    \centering
    \resizebox{\linewidth}{!}{
    \begin{tabular}{l !{\vrule width 1.1pt} cccc}
    \specialrule{1.2pt}{0pt}{0pt}
    \textbf{Setting} & \textbf{OLMo-1B} & \textbf{LLaMA1-7B} & \textbf{Amber-13B} & \textbf{LLaMA1-65B} \\
    \specialrule{0.9pt}{0pt}{0pt}
    Separate C4\&CC & 19.52 & 42.42 & 53.49 & 42.52  \\
    \rowcolor{gray!10} Merge C4\&CC & 94.46 & 95.14 & 78.87 & 94.26  \\
    \specialrule{1.2pt}{0pt}{0pt}
    \end{tabular}
    }
    \vspace{-0.25em}
    \caption{ Similar Dataset Category Merging}
    \vspace{-0.55em}
    \label{tab:c4_cc_merge}
\end{subtable}
\vspace{-0.5em}
\caption{More ablation Studies. Panel (a) confirms the necessity of our inverse correction module. Panel (b) illustrates that merging semantically indistinguishable sources is critical to avoiding ill-conditioned inversion and ensuring stable estimation.}
\label{inverse_and_merging}
\vspace{-.6em}
\end{table}

\begin{table*}[t]
\centering
\small
\resizebox{\textwidth}{!}{
\begin{tabular}{lccccccc}
\toprule
Model & Web & GitHub & Wikipedia & Books & ArXiv & StackExchange & Overlap Accuracy (\%) \\
\midrule
OLMo-1B & 81.10$\rightarrow$83.99 & 13.40$\rightarrow$12.89 & 0.10$\rightarrow$2.04 & 0.20$\rightarrow$0.91 & 2.30$\rightarrow$0.09 & 2.90$\rightarrow$0.08 & 94.46 \\
OLMo-2 & 95.50$\rightarrow$84.99 & 2.10$\rightarrow$7.05 & 0.10$\rightarrow$2.81 & 1.50$\rightarrow$2.02 & 0.50$\rightarrow$0.19 & 0.30$\rightarrow$2.94 & 89.18 \\
OLMo-3 (held-out) & 76.88$\rightarrow$75.37 & 7.06$\rightarrow$12.38 & 0.04$\rightarrow$8.31 & 12.60$\rightarrow$3.44 & 0.82$\rightarrow$0.09 & 2.60$\rightarrow$0.40 & 86.41 \\
\bottomrule
\end{tabular}}
\caption{Temporal held-out generalization on the OLMo family. Each cell reports GroundTruth$\rightarrow$Prediction (\%). All evaluation choices are fixed before testing OLMo-3.}
\label{tab:olmo_temporal_heldout}
\end{table*}

\begin{table}[t]
\centering
\small
\begin{tabular}{lc}
\toprule
GPT-2 Model & Overlap Accuracy (\%) \\
\midrule
gpt2\_balanced    & 75.62 \\
gpt2\_book\_heavy & 50.15 \\
gpt2\_web\_heavy  & 87.53 \\
\bottomrule
\end{tabular}
\caption{Controlled GPT-2 sandbox results. LLMSurgeon is applied post-hoc without any tuning per model or mixture.}
\label{tab:gpt2_controlled_acc}
\end{table}

\noindent\textbf{Robustness of Sampling Text.} To evaluate the robustness of \texttt{LLMSurgeon} to the sampling process, we examine six distinct prompt styles in Table~\ref{Prompts Style}. We find that \textbf{Neutral} sampling exhibits the highest robustness across general-purpose models, consistently maintaining top-tier performance (e.g., 95.14\% on LLaMA-7B) where other styles fluctuate significantly. For instance, while \textbf{Expository} prompts achieve high accuracy on Amber-13B, they fail catastrophically on OLMo-1B (22.71\%), indicating that strong stylistic biasing can distort the generated distribution $q_\pi(x)$ away from the latent training prior, leading to unreliable estimation. Given this trade-off, we adopt Neutral sampling as the default strategy for general auditing tasks to maximize estimation stability and robustness across diverse models.

\noindent\textbf{Effect of Inverse Bias Correction.}
A core contribution of \texttt{LLMSurgeon} is the use of the confusion matrix $C$ to rectify the proxy classifier's outputs. To quantify this gain, we compare our full method against the \textit{w/o Inverse Correction} baseline in Sec.~\ref{no_inverse_baseline}.
Table~\ref{inverse_and_merging}(a) shows that the inverse correction consistently improves estimation fidelity.
For example, on StarCoder, applying the correction boosts accuracy to a relative improvement of nearly 15\%. This confirms that Eq.~\ref{eq:optimization} effectively disentangles the classifier's systematic confusion, sharpening the final distribution estimate.

\noindent\textbf{Effect of Domains Pre-definition.}
DMS relies on the premise that the target domains are linguistically distinguishable. We investigate this boundary by attempting to separate \textit{C4} from \textit{Common Crawl} in the LLaMA training mixture analysis.
As shown in Table~\ref{inverse_and_merging}(b), treating them as separate labels causes a catastrophic performance drop from 95.14\% to 42.42\%.
This occurs because C4 is merely a filtered subset of Common Crawl; they share the same underlying semantic distribution, making $p(x|\text{C4}) \approx p(x|\text{CC})$. Consequently, the classifier cannot distinguish them, leading to an unstable estimation.
Merging these semantically identical sources restores accuracy to near-perfect levels, validating our strategy of grouping overlapping sources into distinct semantic clusters.

\noindent\textbf{Correlation Analysis.}
To validate the robustness of inversion mechanism, we analyze  relationship between domain classification accuracy and  final mixture recovery. As illustrated in Figure~\ref{class_vs_est}, we observe a strong positive correlation across all model families, with Pearson correlation coefficients consistently exceeding $r=0.85$. The trajectory of the curves demonstrates that as models converge during pretraining, their generated domains become linguistically more distinct. This increased separability reduces the condition number of the confusion matrix, thereby allowing \texttt{LLMSurgeon} to recover the ground-truth mixture with significantly higher precision in later checkpoints.

\subsection{Controlled and Held-Out Generalization}
To address concerns about leakage and protocol overfitting, we conduct two complementary evaluations. In a controlled GPT-2 sandbox with a fixed 7-domain taxonomy and unseen mixtures, as shown in Table~\ref{tab:gpt2_controlled_acc}, LLMSurgeon is applied fully post-hoc using only a frozen proxy classifier and its confusion operator, achieving strong recovery on the Balanced and Web-heavy settings (75.62\% and 87.53\%), while performing worse on the Book-heavy setting (50.15\%) due to stronger semantic overlap and a less well-conditioned inverse problem. We further test protocol-level generalization by fixing all evaluation choices on earlier OLMo releases and transferring them without retuning to OLMo-3, where LLMSurgeon still attains high overlap accuracy (86.41\%) and accurately recovers the dominant Web component (76.88$\rightarrow$75.37). Overall, these results suggest that LLMSurgeon generalizes beyond the reported model set, and that its main limitation arises from domain overlap rather than calibration leakage.

\subsection{Safety Auditing Triage via Toxic Injection}
To demonstrate practical utility of DMS, we conduct a controlled toxic-injection study by training GPT-2 on a fixed 7-domain mixture and replacing 5\%, 10\%, or 20\% of training tokens with RealToxicityPrompts~\citep{gehman2020realtoxicityprompts}, while keeping  total token budget ($\sim$12B) constant. Using an 8-class taxonomy (the original 7 domains plus \textit{Toxic}) without model-specific retuning, LLMSurgeon recovers a monotonically increasing toxic mass estimate  with small absolute errors as shown in Table~\ref{tab:toxic_triage}. This suggests LLMSurgeon can provide a low-cost training-exposure signal for safety triage, helping prioritize checkpoints for expensive red-teaming or human review, while complementing rather than replacing output-based toxicity evaluation.

\begin{table}[t]
\centering
\setlength{\tabcolsep}{2.5pt}
\resizebox{\linewidth}{!}{
\begin{tabular}{lcccc}
\toprule
Model & GT Toxic (\%) & Estimated Toxic (\%)  & Toxic Est. Accuracy (\%) \\
\midrule
5\% Toxic GPT-2  & 5.00  & 7.90   & 97.10 \\
10\% Toxic GPT-2 & 10.00 & 12.00  & 98.00 \\
20\% Toxic GPT-2 & 20.00 & 22.73  & 97.27 \\
\bottomrule
\end{tabular}}
\caption{Safety-auditing triage via toxic injection. LLMSurgeon recovers a monotonic estimate of toxic training mass under controlled GPT-2 pretraining.}
\label{tab:toxic_triage}
\end{table}

\section{Conclusion}
\label{sec:conclusion}

We introduced \textbf{\texttt{LLMSurgeon}}, a principled framework for diagnosing implicit training data composition of Large Language Models.
By formalizing \textbf{\texttt{Data Mixture Surgery (DMS)}}, we demonstrate that it is possible to recover  latent domain prior from generated texts.
We further established \textbf{\texttt{LLMScan}}, a comprehensive benchmark built on open-source models with verifiable ground truths, exposing  limitations of traditional membership inference aggregation in this macroscopic setting.
As AI development becomes increasingly opaque, \texttt{LLMSurgeon} establishes a vital, post-hoc mechanism for enforcing data transparency and accountability without relying on voluntary disclosure.

\section{Limitations and Future Work}
\label{sec:limitations}

While \texttt{LLMSurgeon} provides an effective framework for auditing pretraining data, several limitations suggest directions for future research.

First, our method operates under the label-shift assumption, implying that neutral prompts elicit a generation distribution faithfully reflecting the model's training prior. This relationship may be distorted in models that have undergone extensive post-training alignment (e.g., RLHF or instruction tuning), which can shift the output distribution away from the original data mixture. Future work could investigate "inverse-alignment" techniques to disentangle the base distribution from alignment artifacts. 
Second, our method relies on a closed-world assumption defined by the auxiliary classifier's fixed taxonomy. Consequently, it cannot discover novel domains outside these $K$ categories.

Finally, as observed in our fine-grained analysis (Figure~\ref{grain}), the estimation accuracy is inherently bounded by the semantic separability of the domains. Highly overlapping categories (e.g., distinguishing between C and C++ code) result in dense, ill-conditioned confusion matrices that challenge the stability of the linear inversion. 
Addressing this resolution limit perhaps through hierarchical inference strategies or non-linear transport methods remains a promising direction for future exploration in pretraining data auditing.
Extending our framework to broader model families, multilingual corpora would further validate its generality.

\section*{Ethics Statement}
The primary goal of this work is to advance transparency and accountability in the development of large language models by enabling the external auditing of opaque pretraining data mixture. By recovering the implicit composition of training corpora, \texttt{LLMSurgeon} facilitates the identification of potential biases, copyright infringements, and the over-representation of specific viewpoints, thereby empowering researchers and regulators to better understand foundation models. However, we acknowledge that this technology could be repurposed to reverse-engineer proprietary dataset curation strategies or identify vulnerabilities in specific models by revealing their lack of exposure to certain domains. We believe that the scientific and social benefits of open and verifiable model auditing outweigh these risks, and we emphasize that our method operates at the distributional level rather than extracting individual training examples or private data.

\noindent Regarding the using of AI, we just use generative models for writing assistance and coding drafting.

\section*{Acknowledgements}

This work is supported by the MBZUAI-WIS Joint Program for Artificial Intelligence Research.

\bibliography{custom}

@article{qwen3,
  title={Qwen3 technical report},
  author={Yang, An and Li, Anfeng and Yang, Baosong and Zhang, Beichen and Hui, Binyuan and Zheng, Bo and Yu, Bowen and Gao, Chang and Huang, Chengen and Lv, Chenxu and others},
  journal={arXiv preprint arXiv:2505.09388},
  year={2025}
}

@article{deepseek3.2,
  title={Deepseek-v3. 2: Pushing the frontier of open large language models},
  author={Liu, Aixin and Mei, Aoxue and Lin, Bangcai and Xue, Bing and Wang, Bingxuan and Xu, Bingzheng and Wu, Bochao and Zhang, Bowei and Lin, Chaofan and Dong, Chen and others},
  journal={arXiv preprint arXiv:2512.02556},
  year={2025}
}

@article{gemini3,
  title={Gemini 3 Technical Report},
  author={{Gemini Team}},
  year={2025},
  howpublished={\url{https://deepmind.google/technologies/gemini/}}}

@article{gpt5.2,
  title={Introducing {GPT-5.2}},
  author={{OpenAI}},
  journal={OpenAI Blog},
  year={2025},
  month={dec},
  url={https://openai.com/index/introducing-gpt-5-2/}
}

@inproceedings{shokri2017membership,
  title={Membership inference attacks against machine learning models},
  author={Shokri, Reza and Stronati, Marco and Song, Congzheng and Shmatikov, Vitaly},
  booktitle={2017 IEEE symposium on security and privacy (SP)},
  pages={3-18},
  year={2017}
}

@inproceedings{carlini2021extracting,
  title={Extracting Training Data from Large Language Models},
  author={Carlini, Nicholas and Tramer, Florian and Wallace, Eric and Jagielski, Matthew and Herbert-Voss, Ariel and Lee, Katherine and Roberts, Adam and Brown, Tom and Song, Dawn and Erlingsson, Ulfar and others},
  booktitle={USENIX Security Symposium},
  volume={6},
  year={2021}
}

@inproceedings{abadi2016deep,
  title={Deep learning with differential privacy},
  author={Abadi, Martin and Chu, Andy and Goodfellow, Ian and McMahan, H Brendan and Mironov, Ilya and Talwar, Kunal and Zhang, Li},
  booktitle={Proceedings of the 2016 ACM SIGSAC conference on computer and communications security},
  pages={308--318},
  year={2016}
}

@inproceedings{watson2022importance,
  title={On the Importance of Difficulty Calibration in Membership Inference Attacks},
  author={Watson, Lauren and Guo, Chuan and Cormode, Graham and Sablayrolles, Alexandre},
  booktitle={International Conference on Learning Representations},
  year={2022}
}

@inproceedings{duci,
  title={How much of my dataset did you use? Quantitative Data Usage Inference in Machine Learning},
  author={Tong, Yao and Ye, Jiayuan and Zarifzadeh, Sajjad and Shokri, Reza},
  booktitle={ICLR 2025 Workshop on Navigating and Addressing Data Problems for Foundation Models},
  year={2025}
}

@inproceedings{xie2023doremi,
  title={Doremi: Optimizing data mixtures speeds up language model pretraining},
  author={Xie, Sang Michael and Pham, Hieu and Dong, Xuanyi and Du, Nan and Liu, Hanxiao and Lu, Yifan and Liang, Percy and Le, Quoc V and Ma, Tengyu and Yu, Adams Wei},
  booktitle={NeurIPS},
  year={2023}
}

@inproceedings{chen2023datajuicer,
  title={Data-juicer: A one-stop data processing system for large language models},
  author={Chen, Daoyuan and Huang, Yilun and Ma, Zhijian and Chen, Hesen and Pan, Xuchen and Ge, Ce and Gao, Dawei and Xie, Yuexiang and Liu, Zhaoyang and Gao, Jinyang},
  booktitle={Companion of the 2024 International Conference on Management of Data},
  pages={120--134},
  year={2024}
}

@inproceedings{xie2023dsir,
  title={Data Selection for Language Models via Importance Resampling},
  author={Xie, Sang Michael and Santurkar, Shibani and Ma, Tengyu and Liang, Percy},
  booktitle={Advances in Neural Information Processing Systems (NeurIPS)},
  year={2023}
}

@inproceedings{chen2024skillit,
  title={Skill-It! A Data-Driven Skills Framework for Understanding and Training Language Models},
  author={Chen, Mayee and Roberts, Nicholas and Bhatia, Kush and Olmo, Jartu and Diakonikolas, Jelena and Re, Christopher},
  booktitle={International Conference on Learning Representations (ICLR)},
  year={2024}
}

@inproceedings{carlini2022membership,
  title={Membership inference attacks from first principles},
  author={Carlini, Nicholas and Chien, Steve and Nasr, Milad and Song, Shuang and Terzis, Andreas and Tramer, Florian},
  booktitle={2022 IEEE symposium on security and privacy (SP)},
  pages={1897--1914},
  year={2022},
  organization={IEEE}
}

@inproceedings{yeom2018privacy,
  title={Privacy risk in machine learning: Analyzing the connection to overfitting},
  author={Yeom, Samuel and Giacomelli, Irene and Fredrikson, Matt and Jha, Somesh},
  booktitle={2018 IEEE 31st computer security foundations symposium (CSF)},
  pages={268--282},
  year={2018},
  organization={IEEE}
}

@article{Ref,
  title={TLDR: Extreme summarization of scientific documents},
  author={Cachola, Isabel and Lo, Kyle and Cohan, Arman and Weld, Daniel S},
  journal={arXiv preprint arXiv:2004.15011},
  year={2020}
}

@inproceedings{Zlib,
  title={Extracting training data from large language models},
  author={Carlini, Nicholas and Tramer, Florian and Wallace, Eric and Jagielski, Matthew and Herbert-Voss, Ariel and Lee, Katherine and Roberts, Adam and Brown, Tom and Song, Dawn and Erlingsson, Ulfar and others},
  booktitle={30th USENIX security symposium (USENIX Security 21)},
  pages={2633--2650},
  year={2021}
}

@article{mink,
  title={Detecting Pretraining Data from Large Language Models},
  author={Shi, Weijia and Ajith, Anirudh and Xia, Mengzhou and Huang, Yangsibo and Liu, Dennis and Blevins, Terra and Chen, Danqi and Zettlemoyer, Luke},
  journal={arXiv preprint arXiv:2310.16789},
  year={2023}
}

@article{minkpp,
  title={Min-K\%++: Improved Baseline for Detecting Pre-Training Data from Large Language Models},
  author={Zhang, Jingyang and Bao, Yaming and Wen, Deming and Fan, Houqiang and Lin, Wenyuan and Liu, Kai and Zhao, Wayne Xin and Wen, Ji-Rong},
  journal={arXiv preprint arXiv:2404.02936},
  year={2024}
}

@inproceedings{Neighbor,
  title={Membership Inference against Language Models via Neighborhood Comparison},
  author={Mattern, Justus and Shao, Fatemeh and Sablayrolles, Alexandre and Kochems, Pierre},
  booktitle={Findings of the Association for Computational Linguistics: ACL 2023},
  pages={11330--11343},
  year={2023}
}

@article{wang2024pandora,
  title={Pandora's White-Box: Precise Training Data Detection and Extraction in Large Language Models},
  author={Wang, Jeffrey G and Wang, Jason and Li, Marvin and Neel, Seth},
  journal={arXiv preprint arXiv:2402.17012},
  year={2024}
}

@article{recall,
  title={Recall: Membership inference via relative conditional log-likelihoods},
  author={Xie, Roy and Wang, Junlin and Huang, Ruomin and Zhang, Minxing and Ge, Rong and Pei, Jian and Gong, Neil Zhenqiang and Dhingra, Bhuwan},
  journal={arXiv preprint arXiv:2406.15968},
  year={2024}
}

@article{dcpdd,
  title={Pretraining data detection for large language models: A divergence-based calibration method},
  author={Zhang, Weichao and Zhang, Ruqing and Guo, Jiafeng and de Rijke, Maarten and Fan, Yixing and Cheng, Xueqi},
  journal={arXiv preprint arXiv:2409.14781},
  year={2024}
}

@article{shen2023slimpajama,
  title={Slimpajama-dc: Understanding data combinations for llm training},
  author={Shen, Zhiqiang and Tao, Tianhua and Ma, Liqun and Neiswanger, Willie and Liu, Zhengzhong and Wang, Hongyi and Tan, Bowen and Hestness, Joel and Vassilieva, Natalia and Soboleva, Daria and others},
  journal={arXiv preprint arXiv:2309.10818},
  year={2023}
}

@article{gao2020pile,
  title={The Pile: An 800GB Dataset of Diverse Text for Language Modeling},
  author={Gao, Leo and Biderman, Stella and Black, Sid and Golding, Laurence and Hoppe, Travis and Foster, Charles and Phang, Jason and He, Horace and Thite, Anish and Nabeshima, Noa and others},
  journal={arXiv preprint arXiv:2101.00027},
  year={2020}
}

@article{kocetkov2022stack,
  title={The Stack: 3 TB of permissively licensed source code},
  author={Kocetkov, Denis and Li, Raymond and Allal, Loubna Ben and Li, Jia and Mou, Chenghao and Ferrandis, Carlos Mu{\~n}oz and Jernite, Yacine and Mitchell, Margaret and Hughes, Sean and Wolf, Thomas and others},
  journal={arXiv preprint arXiv:2211.15533},
  year={2022}
}

@article{llama1,
  title={Llama: Open and efficient foundation language models},
  author={Touvron, Hugo and Lavril, Thibaut and Izacard, Gautier and Martinet, Xavier and Lachaux, Marie-Anne and Lacroix, Timoth{\'e}e and Rozi{\`e}re, Baptiste and Goyal, Naman and Hambro, Eric and Azhar, Faisal and others},
  journal={arXiv preprint arXiv:2302.13971},
  year={2023}
}

@article{olmo1,
  title={OLMo: Accelerating the Science of Language Models},
  author={Groeneveld, Dirk and Beltagy, Iz and Walsh, Pete and Bhagia, Akshita and Kinney, Rodney and Tafjord, Oyvind and Jha, Ananya Harsh and Ivison, Hamish and Magnusson, Ian and Wang, Yizhong and others},
  journal={arXiv preprint arXiv:2402.00838},
  year={2024}
}

@inproceedings{amber,
  title={LLM360: Towards Fully Transparent Open-Source LLMs},
  author={Liu, Zhengzhong and Qiao, Aurick and Neiswanger, Willie and Wang, Hongyi and Paik, Bowen and Li, Atanc and Li, Gen and Geng, Xinyang and Wang, Renrubin and Sun, Yiran and others},
  booktitle={arXiv preprint arXiv:2312.06550},
  note={Reference for Amber},
  year={2023}
}

@inproceedings{pythia,
  title={Pythia: A Suite for Analyzing Large Language Models Across Training and Scaling},
  author={Biderman, Stella and Schoelkopf, Hailey and Anthony, Quentin and Bradley, Herbie and O'Brien, Kyle and Hallahan, Eric and Khan, Mohammad Aflah and Purohit, Shivanshu and Prashanth, USVSN and Raff, Edward and others},
  booktitle={International Conference on Machine Learning},
  pages={2397--2430},
  year={2023}
}

@article{starcoder,
  title={StarCoder: may the source be with you!},
  author={Li, Raymond and Allal, Loubna Ben and Kocetkov, Denis and Mou, Chenghao and Ferrandis, Carlos Mu{\~n}oz and Santacroce, Sean and Hughes, Sean and Belkada, Younes and others},
  journal={arXiv preprint arXiv:2305.06161},
  year={2023}
}

@article{gptneo,
  title={GPT-Neo: Large Scale Autoregressive Language Modeling with Mesh-Tensorflow},
  author={Black, Sid and Gao, Leo and Wang, Phil and Leahy, Connor and Biderman, Stella},
  year={2021},
  publisher={Zenodo}
}

@inproceedings{soldaini2024dolma,
  title={Dolma: An open corpus of three trillion tokens for language model pretraining research},
  author={Soldaini, Luca and Kinney, Rodney and Bhagia, Akshita and Schwenk, Dustin and Atkinson, David and Authur, Russell and Bogin, Ben and Chandu, Khyathi and Dumas, Jennifer and Elazar, Yanai and others},
  booktitle={Proceedings of the 62nd annual meeting of the association for computational linguistics (volume 1: long papers)},
  pages={15725--15788},
  year={2024}
}

@inproceedings{xiao2024any,
  title={Any-shift prompting for generalization over distributions},
  author={Xiao, Zehao and Shen, Jiayi and Derakhshani, Mohammad Mahdi and Liao, Shengcai and Snoek, Cees GM},
  booktitle={Proceedings of the IEEE/CVF Conference on Computer Vision and Pattern Recognition},
  pages={13849--13860},
  year={2024}
}

@article{kirk2023understanding,
  title={Understanding the effects of rlhf on llm generalisation and diversity},
  author={Kirk, Robert and Mediratta, Ishita and Nalmpantis, Christoforos and Luketina, Jelena and Hambro, Eric and Grefenstette, Edward and Raileanu, Roberta},
  journal={arXiv preprint arXiv:2310.06452},
  year={2023}
}

@article{maini2021dataset,
  title={Dataset inference: Ownership resolution in machine learning},
  author={Maini, Pratyush and Yaghini, Mohammad and Papernot, Nicolas},
  journal={arXiv preprint arXiv:2104.10706},
  year={2021}
}

@article{zhang2024membership,
  title={Membership inference attacks cannot prove that a model was trained on your data},
  author={Zhang, Jie and Das, Debeshee and Kamath, Gautam and Tram{\`e}r, Florian},
  journal={arXiv preprint arXiv:2409.19798},
  year={2024}
}

@inproceedings{meeus2025sok,
  title={Sok: Membership inference attacks on llms are rushing nowhere (and how to fix it)},
  author={Meeus, Matthieu and Shilov, Igor and Jain, Shubham and Faysse, Manuel and Rei, Marek and de Montjoye, Yves-Alexandre},
  booktitle={2025 IEEE Conference on Secure and Trustworthy Machine Learning (SaTML)},
  pages={385--401},
  year={2025},
  organization={IEEE}
}

@article{duan2024membership,
  title={Do membership inference attacks work on large language models?},
  author={Duan, Michael and Suri, Anshuman and Mireshghallah, Niloofar and Min, Sewon and Shi, Weijia and Zettlemoyer, Luke and Tsvetkov, Yulia and Choi, Yejin and Evans, David and Hajishirzi, Hannaneh},
  journal={arXiv preprint arXiv:2402.07841},
  year={2024}
}

@inproceedings{chen2025statistical,
  title={A statistical and multi-perspective revisiting of the membership inference attack in large language models},
  author={Chen, Bowen and Han, Namgi and Miyao, Yusuke},
  booktitle={Proceedings of the 63rd Annual Meeting of the Association for Computational Linguistics (Volume 1: Long Papers)},
  pages={22854--22874},
  year={2025}
}

@inproceedings{gehman2020realtoxicityprompts,
  title={Realtoxicityprompts: Evaluating neural toxic degeneration in language models},
  author={Gehman, Samuel and Gururangan, Suchin and Sap, Maarten and Choi, Yejin and Smith, Noah A},
  booktitle={Findings of the association for computational linguistics: EMNLP 2020},
  pages={3356--3369},
  year={2020}
}

\clearpage
\appendix
\section*{\Large{Appendix}}
\begin{figure*}[htbp] 
    \centering
    \includegraphics[width=0.95\linewidth]{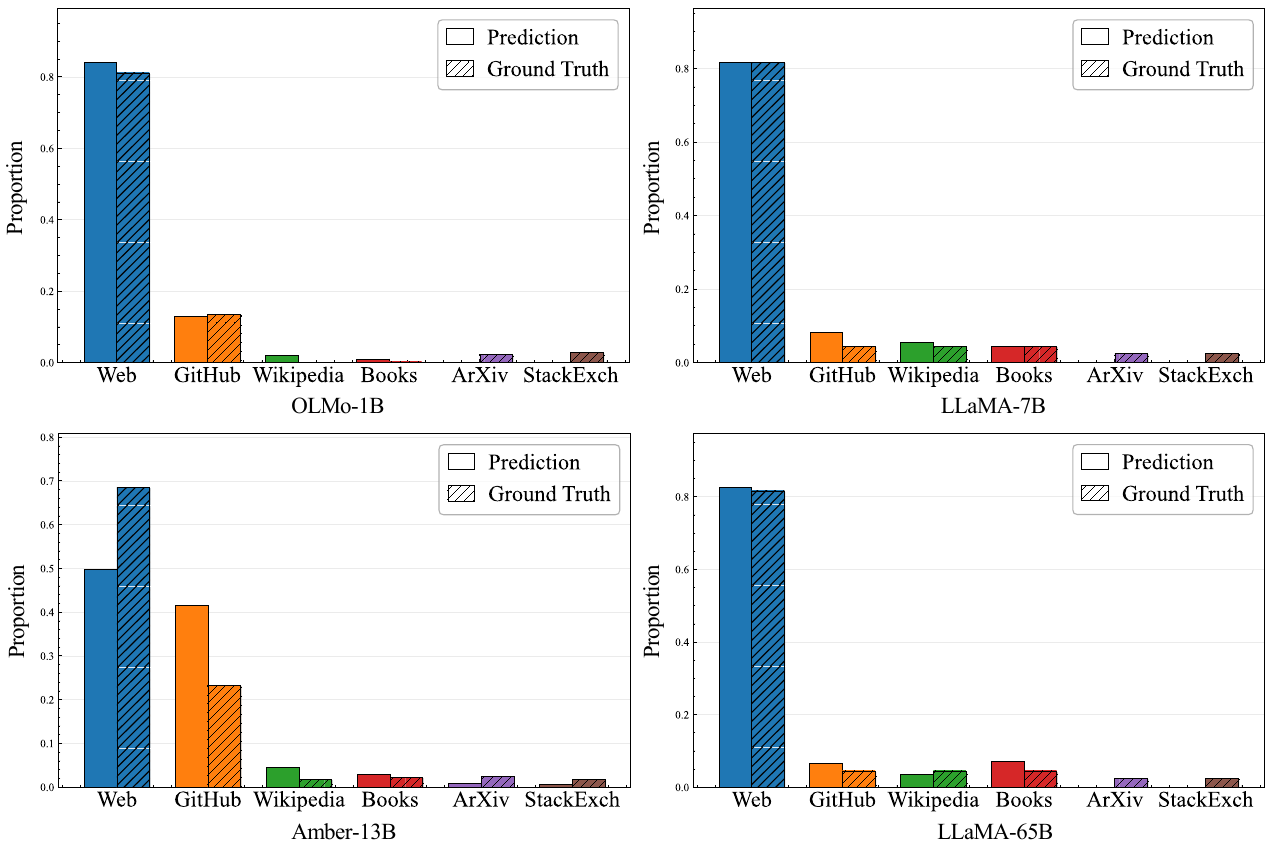}
    \vspace{-0.1in}
    \caption{Predicted versus ground-truth proportions of training data sources for OLMo-1B, Amber-13B, LLaMA1-7B, and LLaMA1-65B across six coarse-grained categories. Solid bars denote model predictions, while hatched bars indicate ground-truth proportions. The figure presents a comparison of detection results across models with different numbers of parameters and across data source categories, providing a coarse-grained view of model predictions relative to the ground truth.}
    \label{appendix:vis_easy}
\end{figure*}

\section{More Experiments Details}

Pretraining Data Mixture Details for the Controlled GPT-2 sandbox Experiment in Table~\ref{tab:gpt2_controlled_acc}.

\begin{table}[h]
\centering
\small
\begin{tabular}{lccc}
\toprule
Domain & Balanced & Book-heavy & Web-heavy \\
\midrule
CommonCrawl   & 20 & 5  & 45 \\
C4            & 20 & 5  & 30 \\
Book          & 20 & 70 & 2  \\
Wikipedia     & 10 & 10 & 7  \\
ArXiv         & 10 & 5  & 3  \\
GitHub        & 10 & 3  & 5  \\
StackExchange & 10 & 2  & 8  \\
\bottomrule
\end{tabular}
\caption{Controlled GPT-2 pretraining mixtures (ground truth, \%).}
\label{tab:gpt2_controlled_mix}
\end{table}

\section{Class-wise Detection Error}

\begin{table*}[t]
    \centering
    \resizebox{0.8\linewidth}{!}{
    \begin{tabular}{cc|cccccc}
        \toprule
        Model Name& &Web& GitHub & Wikipedia & Books & ArXiv & StackExchange \\
        \midrule
        \multirow{3}{*}{OLMo-1B} & Ground Truth & 81.10 & 13.40 & 0.10& 0.20& 2.30& 2.90  \\
         & Prediction & 83.99 & 12.89& 2.04&  0.91&  0.09&  0.08  \\
         & \cellcolor[HTML]{EFEFEF}\textbf{Detection Error} &\cellcolor[HTML]{EFEFEF}2.89 &\cellcolor[HTML]{EFEFEF}0.51 & \cellcolor[HTML]{EFEFEF}1.94 & \cellcolor[HTML]{EFEFEF}0.71 & \cellcolor[HTML]{EFEFEF}2.21 & \cellcolor[HTML]{EFEFEF}2.82 \\
         \midrule

        \multirow{3}{*}{Amber-13B} & Ground Truth & 68.50 & 23.30 & 1.70& 2.30 & 2.50& 1.70  \\
         & Prediction & 49.69 & 41.56& 4.45 &  2.98&  0.78&  0.53  \\
         & \cellcolor[HTML]{EFEFEF}\textbf{Detection Error} &\cellcolor[HTML]{EFEFEF}18.81 &\cellcolor[HTML]{EFEFEF}18.26 & \cellcolor[HTML]{EFEFEF}2.75 & \cellcolor[HTML]{EFEFEF}0.68 & \cellcolor[HTML]{EFEFEF}1.72 & \cellcolor[HTML]{EFEFEF}1.17 \\
        \midrule
        
        \multirow{3}{*}{LLaMA1-7B} & Ground Truth & 81.59 & 4.48 & 4.48 & 4.48 & 2.49 & 2.49  \\
         & Prediction & 81.58 & 8.27& 5.55 & 4.47 & 0.07 & 0.06 \\
         & \cellcolor[HTML]{EFEFEF}\textbf{Detection Error} &\cellcolor[HTML]{EFEFEF}0.01 &\cellcolor[HTML]{EFEFEF}3.79 & \cellcolor[HTML]{EFEFEF}1.07 & \cellcolor[HTML]{EFEFEF}0.01 & \cellcolor[HTML]{EFEFEF}2.42 & \cellcolor[HTML]{EFEFEF}2.43 \\
         \midrule

        \multirow{3}{*}{LLaMA1-65B} & Ground Truth & 81.59 & 4.48 & 4.48 & 4.48 & 2.49 & 2.49 \\
        & Prediction & 82.58 & 6.48 & 3.59 & 7.21 & 0.08 & 0.05  \\
        & \cellcolor[HTML]{EFEFEF}\textbf{Detection Error} &\cellcolor[HTML]{EFEFEF}0.99 &\cellcolor[HTML]{EFEFEF}2.00 & \cellcolor[HTML]{EFEFEF}0.89 & \cellcolor[HTML]{EFEFEF}2.73 & \cellcolor[HTML]{EFEFEF}2.41 & \cellcolor[HTML]{EFEFEF}2.44 \\
         \bottomrule
    \end{tabular}
    }
    \caption{Per-class detection error analysis across six classes for OLMo-1B, Amber-13B, LLaMA1-7B, and LLaMA1-65B, with errors reported as percentage absolute deviations for each category. The table reports the ground-truth proportions, model predictions, and the corresponding absolute detection errors for each class, enabling a detailed per-class comparison of detection results across the evaluated models.}
    \label{tab:appendix-coarse}
\end{table*}

\begin{table*}[htbp]
  \centering
  \begin{subtable}{\linewidth}
    \centering
    \resizebox{\linewidth}{!}{
    \begin{tabular}{cc|ccccccccc}
        \toprule
        Model Name& &Common Crawl& GitHub & Wikipedia & Gutenberg  & ArXiv & StackExchange & PubMed Central& FreeLaw  & USPTO Backgrounds  \\
        \midrule
        \multirow{3}{*}{Pythia-2.8B} & GT  &34.05&8.80&1.77&2.51&10.38&5.95&16.69&7.09&4.23\\
         & Pred & 53.21 &17.09&1.50&0.06&6.01&0.98&3.35&1.16&3.73  \\
         & \cellcolor[HTML]{EFEFEF}\textbf{Error} & \cellcolor[HTML]{EFEFEF} 19.16 & \cellcolor[HTML]{EFEFEF}8.29 &\cellcolor[HTML]{EFEFEF}0.27 & \cellcolor[HTML]{EFEFEF}2.45& \cellcolor[HTML]{EFEFEF}4.37& \cellcolor[HTML]{EFEFEF}4.97 & \cellcolor[HTML]{EFEFEF}13.34& \cellcolor[HTML]{EFEFEF}5.93 &\cellcolor[HTML]{EFEFEF}0.50\\
        \midrule
        \multirow{3}{*}{Pythia-12B} & GT &34.05&8.80&1.77&2.51&10.38&5.95&16.69&7.09&4.23 \\
         & Pred &48.74 & 18.21 &0.97 & 0.07& 8.17&1.11&4.71&0.84&2.71   \\
         & \cellcolor[HTML]{EFEFEF}\textbf{Error} & \cellcolor[HTML]{EFEFEF}14.69 & \cellcolor[HTML]{EFEFEF}9.41&\cellcolor[HTML]{EFEFEF}0.8&\cellcolor[HTML]{EFEFEF}2.44&\cellcolor[HTML]{EFEFEF}2.21&\cellcolor[HTML]{EFEFEF}4.84&\cellcolor[HTML]{EFEFEF}11.98&\cellcolor[HTML]{EFEFEF}6.25&\cellcolor[HTML]{EFEFEF}1.52\\
        \midrule
        \multirow{3}{*}{GPT-Neo-2.7B} & GT &34.05&8.80&1.77&2.51&10.38&5.95&16.69&7.09&4.23 \\
         & Pred&60.5&4.21&1.43&2.67&3.88&0.31&7.36&0.92&6.72\\
         & \cellcolor[HTML]{EFEFEF}\textbf{Error}&\cellcolor[HTML]{EFEFEF}26.45 &\cellcolor[HTML]{EFEFEF} 4.59&\cellcolor[HTML]{EFEFEF}0.34&\cellcolor[HTML]{EFEFEF}0.16&\cellcolor[HTML]{EFEFEF}6.50& \cellcolor[HTML]{EFEFEF}5.64& \cellcolor[HTML]{EFEFEF}9.33&\cellcolor[HTML]{EFEFEF}6.17&\cellcolor[HTML]{EFEFEF}2.49 \\
         \bottomrule
    \end{tabular}
    }
    \caption{Per-class detection error for classes 0-10.}
    
  \end{subtable}

  \vspace{0.8em}
  \begin{subtable}{\linewidth}
    \centering
    \resizebox{\linewidth}{!}{
    \begin{tabular}{cc|ccccccccc}
        \toprule
        Model Name& & PubMed Abstracts &DM Mathematics &Ubuntu IRC&EuroParl&  HackerNews&  PhilPapers &  NIH ExPorter &  Enron Emails \\
        \midrule
        
        \multirow{3}{*}{Pythia-2.8B} & GT &3.56&1.43 &1.02&0.84&0.72&0.44&0.35&0.17 \\
         & Pred  & 0.72 & 0.70& 0.11&0.40& 0.67& 2.08&5.96&2.27 \\
         & \cellcolor[HTML]{EFEFEF}\textbf{Error} & \cellcolor[HTML]{EFEFEF}2.84 & \cellcolor[HTML]{EFEFEF}0.73&\cellcolor[HTML]{EFEFEF}0.91&\cellcolor[HTML]{EFEFEF}0.44&\cellcolor[HTML]{EFEFEF}0.05&\cellcolor[HTML]{EFEFEF}1.64&\cellcolor[HTML]{EFEFEF}5.61&\cellcolor[HTML]{EFEFEF}2.10\\
        \midrule
        
         \multirow{3}{*}{Pythia-12B} & GT &3.56&1.43 &1.02&0.84&0.72&0.44&0.35&0.17 \\
         & Pred & 0.91&1.42&0.14&0.54&0.57&1.55&7.54&1.79  \\
         & \cellcolor[HTML]{EFEFEF}\textbf{Error} & \cellcolor[HTML]{EFEFEF}2.65&\cellcolor[HTML]{EFEFEF}0.01&\cellcolor[HTML]{EFEFEF}0.88&\cellcolor[HTML]{EFEFEF}0.30&\cellcolor[HTML]{EFEFEF}0.15&\cellcolor[HTML]{EFEFEF}1.11&\cellcolor[HTML]{EFEFEF}7.19&\cellcolor[HTML]{EFEFEF}1.62\\
        \midrule

        \multirow{3}{*}{GPT-Neo-2.7B} & GT &3.56&1.43 &1.02&0.84&0.72&0.44&0.35&0.17 \\
         & Pred  &0.68&0.90&0.08&0.16&0.19&1.99&7.17&0.85 \\
         & \cellcolor[HTML]{EFEFEF}\textbf{Error} &\cellcolor[HTML]{EFEFEF}2.88&\cellcolor[HTML]{EFEFEF}0.53&\cellcolor[HTML]{EFEFEF}0.94&\cellcolor[HTML]{EFEFEF}0.68&\cellcolor[HTML]{EFEFEF}0.53&\cellcolor[HTML]{EFEFEF}1.55&\cellcolor[HTML]{EFEFEF}6.82&\cellcolor[HTML]{EFEFEF}0.68\\
         \bottomrule
    \end{tabular}
    }
    \caption{Per-class detection error for classes 10-17.}
    
  \end{subtable}
  \caption{Per-class detection error analysis across 17 classes for models trained on the Pile dataset, with errors reported as percentage absolute deviations for each category.}
  \label{tab:appendix-mid}

\end{table*}

\begin{table*}[htbp]
  \centering
  \begin{subtable}{\linewidth}
    \centering
    \resizebox{0.9\linewidth}{!}{
    \begin{tabular}{c|cccccccccc}
        \toprule
         &ada & agda & alloy & antlr & applescript & assembly & augeas & awk & batchfile & bluespec \\
        \midrule
         GT  & 0.039 & 0.010 & 0.001 & 0.008 & 0.001 & 0.233 & 0.000 & 0.003 & 0.035 & 0.005 \\
         Pred  & 0.230 & 0.073 & 0.111 & 0.042 & 0.458 & 0.603 & 0.076 & 0.227 & 0.379 & 0.036 \\
         \cellcolor[HTML]{EFEFEF}\textbf{Error} & \cellcolor[HTML]{EFEFEF}0.191 & \cellcolor[HTML]{EFEFEF}0.063 & \cellcolor[HTML]{EFEFEF}0.110 & \cellcolor[HTML]{EFEFEF}0.034 & \cellcolor[HTML]{EFEFEF}0.457 & \cellcolor[HTML]{EFEFEF}0.370 & \cellcolor[HTML]{EFEFEF}0.076 & \cellcolor[HTML]{EFEFEF}0.224 & \cellcolor[HTML]{EFEFEF}0.344 & \cellcolor[HTML]{EFEFEF}0.031 \\
         \bottomrule
    \end{tabular}
    }
    \caption{Per-class detection error for classes 0–10 evaluated on StarCoder.}
  \end{subtable}

  \begin{subtable}{\linewidth}
    \centering
    \resizebox{0.9\linewidth}{!}{
    \begin{tabular}{c|cccccccccc}
        \toprule
         &c & c-sharp & clojure & cmake & coffeescript & common-lisp & cpp & css & cuda & dart \\
        \midrule
         GT & 8.082 & 6.697 & 0.069 & 0.068 & 0.095 & 0.210 & 7.336 & 0.450 & 0.084 & 0.549  \\
         Pred & 0.433 & 0.009 & 0.170 & 1.478 & 3.809 & 0.068 & 0.064 & 0.039 & 0.022 & 0.147 \\
         \cellcolor[HTML]{EFEFEF}\textbf{Error} & \cellcolor[HTML]{EFEFEF}7.649 & \cellcolor[HTML]{EFEFEF}6.688 & \cellcolor[HTML]{EFEFEF}0.101 & \cellcolor[HTML]{EFEFEF}1.410 & \cellcolor[HTML]{EFEFEF}3.714 & \cellcolor[HTML]{EFEFEF}0.142 & \cellcolor[HTML]{EFEFEF}7.272 & \cellcolor[HTML]{EFEFEF}0.411 & \cellcolor[HTML]{EFEFEF}0.062 & \cellcolor[HTML]{EFEFEF}0.402 \\
         \bottomrule
    \end{tabular}
    }
    \caption{Per-class detection error for classes 10-20 evaluated on StarCoder.}
  \end{subtable}
  
  \begin{subtable}{\linewidth}
    \centering
    \resizebox{0.9\linewidth}{!}{
    \begin{tabular}{c|cccccccccc}
        \toprule
      &dockerfile & elixir & elm & emacs-lisp & erlang & f-sharp & fortran & glsl & go & groovy\\
        \midrule
         GT & 0.063 & 0.107 & 0.045 & 0.061 & 0.105 & 0.092 & 0.267 & 0.060 & 3.566 & 0.137   \\
         Pred & 0.180 & 2.843 & 0.117 & 0.051 & 0.321 & 0.000 & 1.228 & 0.194 & 0.110 & 0.474 \\
         \cellcolor[HTML]{EFEFEF}\textbf{Error} & \cellcolor[HTML]{EFEFEF}0.117 & \cellcolor[HTML]{EFEFEF}2.736 & \cellcolor[HTML]{EFEFEF}0.072 & \cellcolor[HTML]{EFEFEF}0.010 & \cellcolor[HTML]{EFEFEF}0.216 & \cellcolor[HTML]{EFEFEF}0.092 & \cellcolor[HTML]{EFEFEF}0.961 & \cellcolor[HTML]{EFEFEF}0.134 & \cellcolor[HTML]{EFEFEF}3.456 & \cellcolor[HTML]{EFEFEF}0.337 \\
         \bottomrule
    \end{tabular}
    }
    \caption{Per-class detection error for classes 20-30 evaluated on StarCoder.}
  \end{subtable}

  \begin{subtable}{\linewidth}
    \centering
    \resizebox{0.9\linewidth}{!}{
    \begin{tabular}{c|cccccccccc}
        \toprule
      &haskell & html & idris & isabelle & java & java-server-pages & javascript & json & julia & kotlin \\
        \midrule
         GT & 0.335 & 4.403 & 0.005 & 0.012 & 13.037 & 0.147 & 9.703 & 0.150 & 0.197 & 0.852  \\
         Pred &0.045 & 0.082 & 0.737 & 0.302 & 0.000 & 0.080 & 0.000 & 1.090 & 6.980 & 0.226  \\
         \cellcolor[HTML]{EFEFEF}\textbf{Error} & \cellcolor[HTML]{EFEFEF}0.290 & \cellcolor[HTML]{EFEFEF}4.321 & \cellcolor[HTML]{EFEFEF}0.732 & \cellcolor[HTML]{EFEFEF}0.290 & \cellcolor[HTML]{EFEFEF}13.037 & \cellcolor[HTML]{EFEFEF}0.067 & \cellcolor[HTML]{EFEFEF}9.703 & \cellcolor[HTML]{EFEFEF}0.940 & \cellcolor[HTML]{EFEFEF}6.783 & \cellcolor[HTML]{EFEFEF}0.626 \\
         \bottomrule
    \end{tabular}
    }
    \caption{Per-class detection error for classes 30-40 evaluated on StarCoder.}
  \end{subtable}

  \begin{subtable}{\linewidth}
    \centering
    \resizebox{0.9\linewidth}{!}{
    \begin{tabular}{c|cccccccccc}
        \toprule
      &lean &agda & l-coffeescript & l-haskell & lua & makefile & maple & markdown & mathematica & matlab\\
        \midrule
         GT  & 0.014 & 0.001 & 0.001 & 0.008 & 0.430 & 0.197 & 0.001 & 11.236 & 0.187 & 0.000  \\
         Pred  & 0.705 & 0.000 & 1.594 & 1.869 & 4.263 & 0.058 & 0.790 & 14.348 & 0.308 & 0.904 \\
         \cellcolor[HTML]{EFEFEF}\textbf{Error} & \cellcolor[HTML]{EFEFEF}0.691 & \cellcolor[HTML]{EFEFEF}0.001 & \cellcolor[HTML]{EFEFEF}1.593 & \cellcolor[HTML]{EFEFEF}1.861 & \cellcolor[HTML]{EFEFEF}3.833 & \cellcolor[HTML]{EFEFEF}0.139 & \cellcolor[HTML]{EFEFEF}0.789 & \cellcolor[HTML]{EFEFEF}3.112 & \cellcolor[HTML]{EFEFEF}0.121 & \cellcolor[HTML]{EFEFEF}0.904 \\
         \bottomrule
    \end{tabular}
    }
    \caption{Per-class detection error for classes 40-50 evaluated on StarCoder.}
  \end{subtable}

  \begin{subtable}{\linewidth}
    \centering
    \resizebox{0.9\linewidth}{!}{
    \begin{tabular}{c|cccccccccc}
        \toprule
      &ocaml & pascal & perl & php & powershell & prolog & protocol-buffer & python & r & racket\\
        \midrule
         GT & 0.154 & 0.252 & 0.335 & 9.131 & 0.168 & 0.001 & 0.046 & 9.057 & 0.045 & 0.005  \\
         Pred & 1.256 & 0.188 & 0.453 & 0.155 & 0.688 & 0.819 & 0.049 & 27.029 & 5.590 & 0.106 \\
         \cellcolor[HTML]{EFEFEF}\textbf{Error} & \cellcolor[HTML]{EFEFEF}1.102 & \cellcolor[HTML]{EFEFEF}0.064 & \cellcolor[HTML]{EFEFEF}0.118 & \cellcolor[HTML]{EFEFEF}8.976 & \cellcolor[HTML]{EFEFEF}0.520 & \cellcolor[HTML]{EFEFEF}0.818 & \cellcolor[HTML]{EFEFEF}0.003 & \cellcolor[HTML]{EFEFEF}17.972 & \cellcolor[HTML]{EFEFEF}5.545 & \cellcolor[HTML]{EFEFEF}0.101 \\
         \bottomrule
    \end{tabular}
    }
    \caption{Per-class detection error for classes 50-60 evaluated on StarCoder.}
  \end{subtable}

  \begin{subtable}{\linewidth}
    \centering
    \resizebox{0.9\linewidth}{!}{
    \begin{tabular}{c|cccccccccc}
        \toprule
      &restructuredtext & rmarkdown & ruby & rust & sas & scala & scheme & shell & solidity & sparql \\
        \midrule
         GT & 0.498 & 0.009 & 1.021 & 1.366 & 0.018 & 0.704 & 0.030 & 0.463 & 0.128 & 0.006  \\
         Pred & 3.826 & 2.176 & 1.461 & 0.076 & 0.198 & 0.092 & 0.138 & 0.777 & 0.069 & 0.266\\
         \cellcolor[HTML]{EFEFEF}\textbf{Error} &\cellcolor[HTML]{EFEFEF}3.328 & \cellcolor[HTML]{EFEFEF}2.167 & \cellcolor[HTML]{EFEFEF}0.440 & \cellcolor[HTML]{EFEFEF}1.290 & \cellcolor[HTML]{EFEFEF}0.180 & \cellcolor[HTML]{EFEFEF}0.612 & \cellcolor[HTML]{EFEFEF}0.108 & \cellcolor[HTML]{EFEFEF}0.314 & \cellcolor[HTML]{EFEFEF}0.059 & \cellcolor[HTML]{EFEFEF}0.260  \\
         \bottomrule
    \end{tabular}
    }
    \caption{Per-class detection error for classes 60-70 evaluated on StarCoder.}
  \end{subtable}

    \begin{subtable}{0.9\linewidth}
    \centering
    \resizebox{\linewidth}{!}{
    \begin{tabular}{c|cccccccccc}
        \toprule
      &sql & stan & standard-ml & stata & systemverilog & tcl & tcsh & tex & thrift & typescript\\
        \midrule
         GT  & 1.663 & 0.001 & 0.029 & 0.049 & 0.059 & 0.053 & 0.003 & 0.780 & 0.001 & 3.977 \\
         Pred & 0.655 & 0.042 & 0.699 & 1.178 & 0.309 & 0.355 & 0.186 & 0.152 & 0.109 & 0.302  \\
         \cellcolor[HTML]{EFEFEF}\textbf{Error} & \cellcolor[HTML]{EFEFEF}1.008 & \cellcolor[HTML]{EFEFEF}0.041 & \cellcolor[HTML]{EFEFEF}0.670 & \cellcolor[HTML]{EFEFEF}1.128 & \cellcolor[HTML]{EFEFEF}0.250 & \cellcolor[HTML]{EFEFEF}0.302 & \cellcolor[HTML]{EFEFEF}0.183 & \cellcolor[HTML]{EFEFEF}0.628 & \cellcolor[HTML]{EFEFEF}0.108 & \cellcolor[HTML]{EFEFEF}3.675  \\
         \bottomrule
    \end{tabular}
    }
    \caption{Per-class detection error for classes 70-80 evaluated on StarCoder.}
    \end{subtable}

    \begin{subtable}{0.7\linewidth}
    \centering
    \small
    \resizebox{\linewidth}{!}{
    \begin{tabular}{c|cccccccccc}
        \toprule
      & verilog & vhdl & visual-basic & xslt & yacc & yaml & zig\\
        \midrule
         GT & 0.000 & 0.141 & 0.213 & 0.008 & 0.016 & 0.150 & 0.026  \\
         Pred &0.046 & 0.106 & 0.818 & 0.063 & 0.307 & 0.871 & 0.021 \\
         \cellcolor[HTML]{EFEFEF}\textbf{Error} & \cellcolor[HTML]{EFEFEF}0.046 & \cellcolor[HTML]{EFEFEF}0.035 & \cellcolor[HTML]{EFEFEF}0.605 & \cellcolor[HTML]{EFEFEF}0.055 & \cellcolor[HTML]{EFEFEF}0.291 & \cellcolor[HTML]{EFEFEF}0.721 & \cellcolor[HTML]{EFEFEF}0.005 \\
         
         \bottomrule
    \end{tabular}
    }
    \caption{Per-class detection error for classes 80-87 evaluated on StarCoder.}
  \end{subtable}

  \caption{Per-class detection error analysis across 87 programming languages for the StarCoder model, reporting percentage absolute errors for each language category.}

\end{table*}

\subsection{Coarse-Grained Detection Error}
This section presents a coarse-grained detection error analysis over six high-level data source categories: Web, GitHub, Wikipedia, Books, ArXiv, and StackExchange. These categories represent broad groupings of training data sources and are used to evaluate model performance at an aggregate level. The analysis examines the alignment between model-predicted data source distributions and the corresponding ground-truth distributions under a coarse-grained categorization.

Figure~\ref{appendix:vis_easy} provides a visual comparison of predicted and ground-truth proportions for OLMo-1B, Amber-13B, LLaMA1-7B, and LLaMA1-65B. Solid bars denote model predictions, while hatched bars indicate ground-truth proportions. All values are normalized to sum to one, allowing for direct comparison across categories and models. This visualization offers an overview of detection results prior to detailed numerical inspection.

Table~\ref{tab:appendix-coarse} reports the corresponding numerical results, including ground-truth proportions, model predictions, and absolute detection errors expressed as percentage deviations for each category. Detection errors are computed as the absolute difference between predicted and ground-truth proportions. Overall, the table summarizes coarse-grained detection behavior across models and data sources, with lower errors observed for dominant categories such as Web and GitHub, and comparatively larger deviations for less prevalent categories such as ArXiv and StackExchange.

\subsection{Mid-Grained Detection Error}
To further analyze detection performance beyond coarse-grained categories, this section reports mid-grained detection errors across 17 data source classes derived from the Pile dataset. These classes provide a finer partitioning of training data sources, covering a wider range of domain-specific datasets while remaining more aggregated than individual file types or instances. This mid-grained evaluation offers increased resolution relative to the coarse-grained setting while preserving interpretability.

The results of the mid-grained evaluation are presented in Table~\ref{tab:appendix-mid}, which is divided into two subtables corresponding to class indices 0–10 and 10–17, respectively. For each class and model, we report the ground-truth proportion, the predicted proportion, and the absolute detection error expressed as a percentage. The class indices follow the ordering defined in the Pile dataset specification and are applied consistently across all evaluated models.

Overall, the results indicate larger detection errors for certain broad data sources, such as Common Crawl, while smaller errors are observed for categories such as HackerNews. These observations summarize the mid-grained detection behavior across the evaluated classes without further aggregation.

\subsection{Fine-Grained Detection Error}
We further report a fine-grained detection error analysis for the StarCoder model across 87 programming language categories, where each category corresponds to a distinct programming language. This analysis provides the most detailed view of detection behavior in our evaluation and enables a class-wise inspection of prediction accuracy at the level of individual programming languages.

Due to the large number of categories, the results are organized into multiple subtables, each covering a contiguous range of class indices, collectively spanning all 87 programming language classes. For each language, the tables report the ground-truth proportion, the model-predicted proportion, and the corresponding absolute detection error expressed as a percentage. All values are normalized such that proportions across all classes sum to one.

From the reported results, languages with relatively large ground-truth proportions, such as \textit{Java}, \textit{JavaScript}, \textit{Python}, and \textit{Go}, tend to exhibit larger absolute detection errors. In several cases, the model substantially underestimates these high-frequency languages, leading to large deviations between predicted and ground-truth proportions. For example, Java and JavaScript show zero predicted instances (0\%). reflecting pronounced discrepancies in their predicted distributions.

In contrast, many low-frequency languages, including \textit{Agda}, \textit{Alloy}, \textit{ANTLR}, and \textit{Emacs Lisp}, are associated with comparatively small absolute errors. For these categories, both the ground-truth and predicted proportions are close to zero, resulting in limited absolute deviation despite relative differences. This pattern is consistently observed across multiple subtables.

Additionally, several mid-frequency languages exhibit notable overestimation, such as \textit{Markdown}, \textit{Julia}, and \textit{Elixir}, where predicted proportions exceed the corresponding ground-truth values by a substantial margin. Conversely, languages such as \textit{C}, \textit{C++}, and \textit{Rust} are markedly underestimated, contributing to larger detection errors within their respective class groups.

Overall, the fine-grained results reveal substantial variability in detection accuracy across programming languages, with absolute errors influenced by both the underlying ground-truth frequency of a language and the model’s tendency to over- or under-predict specific language categories. These detailed per-language results serve as a supplementary reference for understanding class-wise detection behavior at the finest granularity.

\end{document}